\documentclass[11pt]{article}
\usepackage[preprint]{acl}
\hypersetup{hyperfootnotes=false}
\usepackage{times}
\usepackage{latexsym}
\usepackage[T1]{fontenc}
\usepackage[utf8]{inputenc}
\usepackage{amsmath,amssymb,amsfonts}
\usepackage{graphicx,booktabs,multirow}
\usepackage{eso-pic}
\usepackage{microtype,inconsolata}
\usepackage{placeins,subcaption,tabularx}
\DeclareUnicodeCharacter{2212}{\ensuremath{-}}
\title{PAT: A Personality-Driven Augmentation and Transfer Learning Framework for Depression Detection on Social Media}
\def\PATWithAuthors{1}
\newcommand{\printcredits}{\section*{CRediT authorship contribution statement}\textbf{Shengwu Zheng :} Conceptualization, Methodology, Software, Writing, Investigation .~ \textbf{Yuzhe Ding :} Methodology, Writing .~ \textbf{Xiaohui Liang :} Conceptualization, Methodology, Supervision, Funding acquisition .~ \textbf{Fei Li :} Conceptualization, Methodology, Supervision .~ \textbf{Chong Teng :} Conceptualization, Methodology, Supervision .~ \textbf{Donghong Ji :} Conceptualization, Methodology, Supervision.}
\author{
\textbf{Shengwu Zheng}\textsuperscript{1} \quad \textbf{Yuzhe Ding}\textsuperscript{1} \quad \textbf{Xiaohui Liang}\textsuperscript{2} \\
\textbf{Fei Li}\textsuperscript{1} \quad \textbf{Chong Teng}\textsuperscript{1} \quad \textbf{Donghong Ji}\textsuperscript{1}\thanks{Corresponding author: dhji@whu.edu.cn.} \\
\normalfont\small\textsuperscript{1}School of Cyber Science and Engineering, Wuhan University \\
\normalfont\small\textsuperscript{2}School of Public Health, Wuhan University \\
\normalfont\small\texttt{zhengshengwu@whu.edu.cn} \quad \texttt{yuzheding@whu.edu.cn} \\
\normalfont\small\texttt{xhliang@whu.edu.cn} \quad \texttt{lifei\_csnlp@whu.edu.cn} \\
\normalfont\small\texttt{tengchong@whu.edu.cn} \quad \texttt{dhji@whu.edu.cn} \\
}
\AddToShipoutPictureFG*{\AtPageLowerLeft{\raisebox{18pt}{\hspace*{\dimexpr1in+\oddsidemargin\relax}\makebox[\textwidth][c]{\fontsize{6.5}{8}\selectfont\rmfamily This work has been submitted to the IEEE for possible publication. Copyright may be transferred without notice, after which this version may no longer be accessible.}}}}
\begin{document}
\maketitle
\begin{abstract}

Depression is a prevalent mental disorder that affects millions worldwide, with early detection crucial for effective intervention. 
While existing methods have achieved promising results in automated depression detection, they face two key limitations: (1) noise in post screening: symptom-based semantic screening is easily misled by superficial lexical patterns, thereby introducing false-positive posts; (2) weak optimization of the post encoder: relying solely on coarse user-level supervision signals prevents models from capturing depressive cues within individual posts.
To address these limitations, we propose PAT, a \textbf{P}ersonality-driven \textbf{A}ugmentation and \textbf{T}ransfer learning framework. 
In the first phase, PAT optimizes a post-level encoder via a fine-grained contrastive learning objective, utilizing LLM-based data augmentation driven by classic personality theories to alleviate data scarcity. 
In the second phase, the trained post-level scorer is deployed for high-quality post filtering, while its encoder parameters are seamlessly transferred for user-level tasks.
Extensive experiments demonstrate that PAT significantly outperforms existing methods. Furthermore, PAT delivers multi-granular interpretability by tracing post-level mental-state trajectories and highlighting key symptoms, thereby offering reliable diagnostic evidence for clinical practice.

\end{abstract}

\section{Introduction} \label{sec:intro}
Depression represents a prevalent and debilitating mental disorder in modern society. According to the World Health Organization (WHO)\footnote{\url{https://www.who.int/news-room/fact-sheets/detail/depression}}, depression affected approximately 332 million people worldwide in 2021, corresponding to 4.0\% of the world’s population.
Its core symptoms include sadness, pessimism, self-dislike and other affective disturbances, which can lead to suicidal behavior in severe cases~\cite{beck1996manual}. 
Early detection and intervention are critical for mitigating disease progression and preventing these adverse outcomes.

\begin{figure}[tbp]
\centering
  \includegraphics[width=0.95\linewidth]{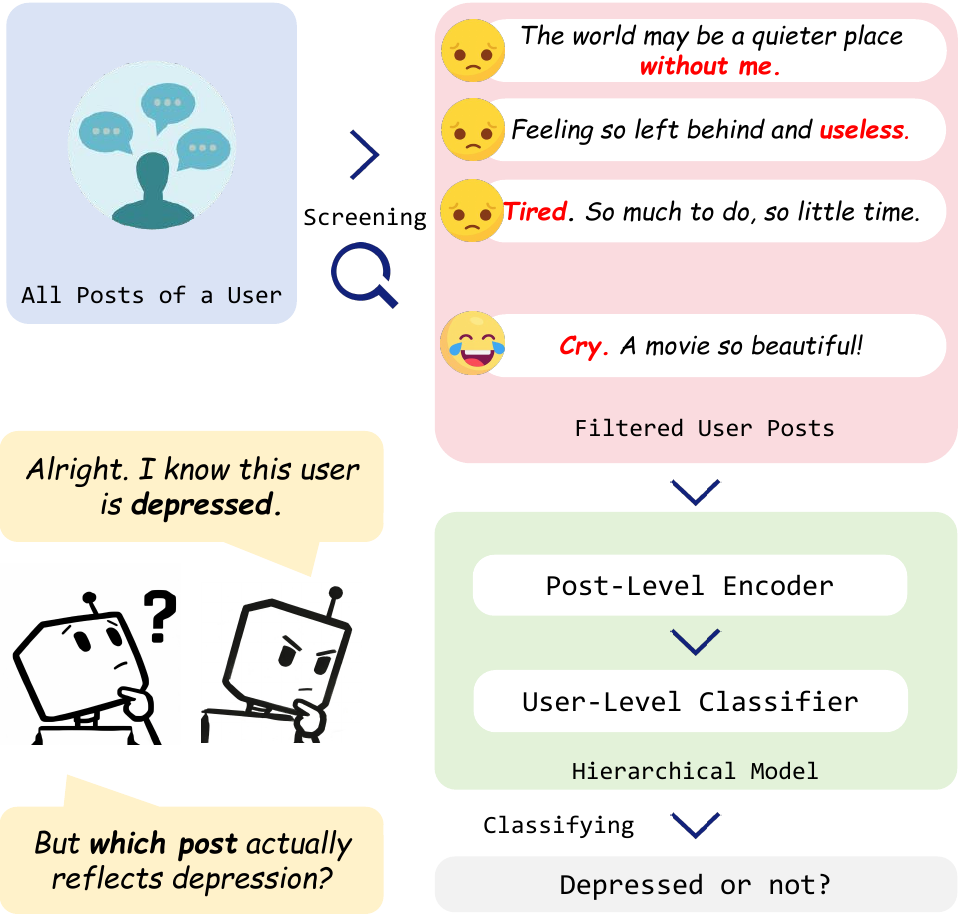}
  \caption{A typical Screen-and-Classify framework for social media user-level depression detection. The hierarchical model is trained solely on binary supervision (depressed or non-depressed), which prevents the post encoder from capturing fine-grained depressive cues in each post and thus limits its optimization. Moreover, the screening step may introduce false positives. For instance, the word “cry” in the fourth sentence of the figure expresses a positive emotion, misleading the model.}
  \label{fig:intro}
\end{figure}

With the popularization of social networks, a growing number of users are willing to express their real emotions and feelings online, providing extensive data to develop automated tools for depression detection~\cite{ding2020depression,wang-etal-2024-esdm,tahir2025depression}. Within this domain, the primary task is to predict a user's depressive status according to their posts, commonly termed user-level depression detection. 
A leading approach for this task is the Screen-and-Classify framework~\cite{zhang2022psychiatric,liu2024depression,wang-etal-2025-end}, as illustrated in Figure~\ref{fig:intro}. 
This framework comprises two steps: (1) Post Screening: retrieving posts related to depression using symptom descriptions as queries; and (2) Hierarchical Classification: encoding the posts and aggregating embeddings for classification. 

However, both steps in this framework face bottlenecks: First, post screening may introduce false positives that contain symptom keywords but express positive meanings. Second, the optimization relies solely on sparse user-level supervision, providing insufficient guidance for the post encoder. 
Besides, these methods only output user-level predictions with limited explanations, lacking the evidence needed for clinical interpretation.
Recent efforts have attempted to address this gap using Large Language Models (LLMs) for post-level information extraction and explanation generation~\cite{lan2025depression,zheng2024cascade}.
Although promising, LLM-based approaches face their own constraints: the risk of hallucination and high computational costs that hinder real-world deployment~\cite{zhang-etal-2025-explainable,ravenda-etal-2025-llms}.

\begin{figure*}[!t]
\centering
  \includegraphics[width=0.90\linewidth]{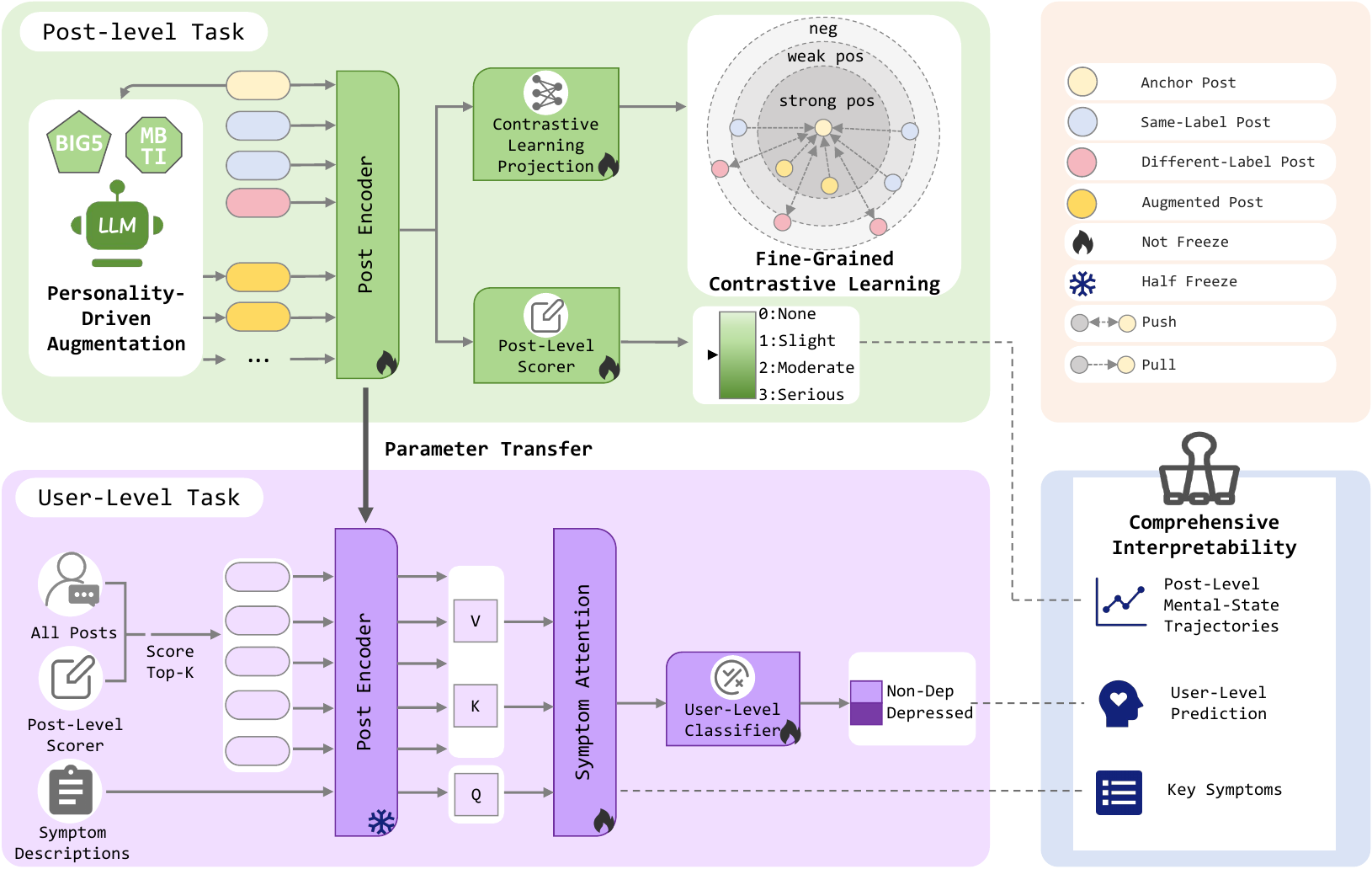}
  \caption {An overview of the PAT framework: it first performs post-level optimization and then transfers the post scorer's encoder to the user-level task for screening and encoding. For post-level optimization, PAT introduces \textbf{Personality-Driven Augmentation} and novel \textbf{Fine-grained Contrastive Learning}. Unlike standard supervised contrastive learning, it distinguishes three data types derived from an anchor post: (1) Augmented Posts, generated under personality guidance, serving as strong positives; (2) Same-Label Posts, which share the ground-truth label, serving as weak positives; and (3) Different-Label Posts, which have different labels, serving as negatives.}
  
  \label{fig:method}
\end{figure*}

To address these limitations, we propose a two-phase \textbf{P}ersonality-driven \textbf{A}ugmentation and \textbf{T}ransfer learning framework (PAT), whose overall structure is shown in Figure~\ref{fig:method}.
In the first phase, PAT trains a post-level scorer to evaluate depressive severity, which is subsequently leveraged to perform score-based filtering on user-level inputs while its encoder component is transferred for user-level training. To alleviate the inherent scarcity of post-level data, we draw inspiration from psycholinguistic findings that personality drives distinct linguistic manifestations~\cite{koutsoumpis2022kernel}. Specifically, we employ LLMs for personality-driven data augmentation based on personality theories, including the Big Five~\cite{costa2008revised} and MBTI~\cite{jung2016psychological}, thus expanding the diversity of data in various expression styles. Furthermore, a fine-grained contrastive learning objective is designed to effectively utilize both the augmented and labeled samples for robust representation learning.
Beyond performance enhancements, PAT distinguishes itself by providing multi-granular interpretability: it delivers user-level predictions, traces post-level mental-state trajectories, and highlights key symptoms, thus offering comprehensive diagnostic evidence. 
In summary, the contributions of this paper are threefold:
\begin{itemize}
    \item We propose a novel transfer learning framework, PAT. Extensive experiments on multiple benchmarks demonstrate that PAT significantly outperforms existing strong baselines and can serve as a plug-and-play component to enhance other methods. 
    
    \item We design a personality-driven data augmentation method with a corresponding fine-grained contrastive learning algorithm. Experiments demonstrate that these methods effectively optimize the embedding space of the post encoder, while revealing the limitations of conventional supervised contrastive learning.
    
    \item We build a comprehensive, interpretable analysis system. Beyond user-level depression detection, PAT traces post-level mental-state trajectories and highlights key symptoms, offering insights that improve practical utility and trustworthiness.
\end{itemize} 

\section{Related Work}
\subsection{User-Level Depression Detection}
User-Level Depression Prediction aims to assess depression risk using a user's social media posts.
Early researchers explored diverse machine learning techniques~\cite{ding2020depression,vasha2023depression,hossain2021social}.
With the rapid advancement of deep learning, various neural networks have become the dominant approach for this task, such as Convolutional Neural Networks~\cite{yates2017depression,lin2020sensemood,narayanan2022detection} and Recurrent Neural Networks~\cite{kour2022hybrid,gamaarachchige2019multi}. \citet{zhang2022psychiatric} proposed a post screening method based on psychological scales and adopted a hierarchical model that sequentially conducts post encoding and user classifying, which has since become a mainstream approach in depression detection.
Following this paradigm, \citet{liu2024depression} introduced capsule networks to model features of different symptoms. 
\citet{wang-etal-2025-end} further improved the screening method by making it end-to-end trainable.
However, these methods typically rely on user-level binary labels for the whole training. This makes it difficult for the models to obtain specific supervision signals at the post level.

Several research efforts have attempted to address this limitation.
\citet{perez-etal-2023-semantic} designed a semantic retrieval framework that performs detection via K-nearest neighbor voting, which is non-parametric.
\citet{nguyen2022improving} trained a separate questionnaire model with a semi-supervised dataset, which resulted in noisy labels. 
\citet{ahmed2022eandc} proposed a curriculum annotation method to improve data quality.
\citet{agarwal2024analyzing} leveraged medical experts to annotate symptom-level scores, but this approach is costly and suffers from label sparsity because certain symptoms are rarely mentioned in personal posts due to privacy concerns. 

\subsection{Transfer Learning}
The core of transfer learning is to leverage knowledge from a data-rich source domain to alleviate data scarcity in a target domain~\cite{pan2009survey}.
In mental-health research,~\citet{ji2022mentalbert} and~\citet{zhai2024chinese} pre-trained a domain-specialized Mental-BERT in the psychological corpora.
\citet{wu2023exploring} transferred knowledge from sentiment classification to depression detection via distillation.
These methods perform pre-training or transfer on domain-related corpora or auxiliary tasks, whose alignment with the target task remains suboptimal. 
In this work, we propose using post-level depression detection as the source task and transfer the post encoder to the user-level depression detection model. This design is expected to achieve a tighter knowledge transfer by directly bridging the gap between the two closely related levels of depression detection.

\section{Proposed Methodology}
The architecture of the proposed PAT framework is shown in Figure~\ref{fig:method}. The framework follows a two-stage design, comprising a post-level task and a user-level task of depression detection. The depression-level scorer trained in the post-level task is deployed for user-level post screening, while its encoder is transferred to the user-level task.

\subsection{Post-Level Depression Detection}

\paragraph{Problem Definition:}
The input is a single social media post denoted as $p$. The objective is to formulate a regression task that estimates a depression-intensity score $r\in[0,3]$, where a higher value indicates increased severity at four progressive levels: None, Slight, Moderate, and Serious.

\subsubsection{Personality-Driven Augmentation}
We employ psychological personality theories to guide a large language model (LLM) in generating augmented samples that reflect distinct personality tendencies, thereby mitigating the scarcity of post-level annotated data.
Specifically, we adopt two mainstream personality theories: (1) the Big Five: taking the high and low tendencies of each of the five traits (Extraversion, Agreeableness, Conscientiousness, Neuroticism, Openness), resulting in ten categories; 
(2) the MBTI (Jungian eight-function): directly using its eight cognitive-function dimensions including Extroverted, Introverted, Sensing, Intuitive, Thinking, Feeling, Judging, and Perceiving. 
We implement the augmentation through a two-step procedure. First, the LLM analyzes the original post with reference to the 21 symptom dimensions of BDI-II, identifies the depression-related symptoms supported by the text, and estimates the severity of each identified symptom. The analysis is returned as a list of symptom--severity pairs. Second, the original post and the symptom--severity pairs are jointly provided to the LLM to generate augmented sentences. We constrain the LLM to preserve the original depressive-symptom semantics while accentuating the linguistic style associated with the specified personality trait. This approach achieves style-diverse and semantically controlled text augmentation. The prompts for both steps are shown in Appendix~\ref{sec:prompt}.

After data augmentation, we perform post-level model training. First, a pretrained encoder $M_{post}$ encodes the post $p_i$ into a text embedding. Then, we utilize a two-layer linear regression head to produce an unnormalized score $\tilde{r}_i$. The Mean Squared Error (MSE) loss is computed as:

\begin{align}
    \hat{r}_i &= 3 \cdot \text{Sigmoid}(\tilde{r}_i) \\
    \mathcal{L}_{\text{mse}} &= \frac{1}{N} \sum_{i=1}^{N} (r_i - \hat{r}_i)^2
    \label{eq:post-level loss}
\end{align}

\subsubsection{Fine-Grained Contrastive Learning}
In addition to the classification objective, we introduce fine-grained contrastive learning to fully leverage augmented data for optimizing the embedding space.
The core idea of contrastive learning is to pull positive (same-class) samples closer and push negative (different-class) samples apart.
Our proposed fine-grained contrastive learning distinguishes two types of positive samples: strong positives (augmented posts) and weak positives (same-label posts), as illustrated in Figure~\ref{fig:method}.

Specifically, for a post embedding $\boldsymbol{h}_
i$ in a data batch, we transform it via a projection head $\Phi$ into a contrastive embedding $\boldsymbol{z}
_i$. Correspondingly, we obtain the set of strong-positive contrastive embeddings $\tilde Z=\{\tilde{\boldsymbol{z}}_1, ...,\tilde{\boldsymbol{z}}_m\}$ and the set of weak-positive embeddings 
$\dot Z=\{\dot{\boldsymbol z}_1, ...,\dot{\boldsymbol z}_t\}$. The fine-grained contrastive loss is formulated as:

\begin{equation}
\begin{split}
    \mathcal{L}_{cl} ={}&\frac{1}{B}\sum_{i \in I} \frac {-1} {|\Gamma(i)|}\sum_ {j\in\Gamma(i)} 
	w_{ij} \cdot \\
    &\log	\frac 	{e ^ {\text{sim}(\boldsymbol{z}_i,\boldsymbol{z}_j)/\tau}}
            	{\sum_{k \in I \setminus i}  e^{\text{sim}(\boldsymbol{z}_i,\boldsymbol{z}_k)/\tau}} 
\end{split}
\end{equation}

where $B$ denotes the batch size, $I=\{1,2,...B\}$ indexes all samples in a batch, $I\setminus i$ denotes the index set of the batch excluding the $i$-th sample, $\Gamma(i)$ represents the set of indices of all positive samples (strong or weak) for the $i$-th sample, $\text{sim}(\boldsymbol{z}_i,\boldsymbol{z}_j)$
is the cosine similarity between the anchor sample $\boldsymbol{z}_i$ and one of its positive samples $\boldsymbol{z}_j$. $\tau$ is the temperature hyper-parameter.
$w_{ij}$ is the weight that is set to 1 for strong positive samples, and dynamically varies between $w_{low}$ and 1 for weak positive samples, depending on its distance to the anchor sample. We define the distance between samples as follows:

\begin{equation}
    d_{ij} = 1-\text{sim}(\boldsymbol{z}_i,\boldsymbol{z}_j)
\end{equation}

The average distance from strong positive samples $\tilde{Z}$ to the anchor sample $\boldsymbol{z}_i$ is calculated as follows: 

\begin{equation}
    \overline{d}_i=\frac{1}{|\tilde Z|}\sum_{\tilde {\boldsymbol{z}} \in \tilde Z}(1-\text{sim}(\boldsymbol{z}_i,\tilde {\boldsymbol{z}}))
\end{equation}

For a weak positive sample $\boldsymbol{z}_j$, its weight decays exponentially with the distance: 

\begingroup
\renewcommand{\arraystretch}{1.2}
\begin{equation}
w_{ij} = \left\{\begin{array}{@{}l@{\hspace{0.05em}}l}
w_{low}+(1-w_{low}) e^{-\gamma \frac {d_{ij}-\overline d_i}{d_{ij}}} & \text{if } d_{ij} > \overline d_i
\\
1 & \text{if } d_{ij} \leq \overline d_i 
\end{array}\right.
\end{equation}
\endgroup

where $\gamma$ is a hyper-parameter and $w_{low}$ is the decay lower bound. 

Finally, let the weighted coefficient for $\mathcal{L}_{cl}$ be $\alpha$, and the overall loss is:

\begin{equation}
    \mathcal{L}_{total} = \mathcal{L}_{mse}+\alpha \mathcal{L}_{cl}
\end{equation}

\subsection{User-Level Depression Detection}

\paragraph{Problem Definition:}
The input is a set of social media posts from a user, denoted as $P=\{p_1,...,p_n\}$ where $p_i$ is the $i$-th post. The objective is to predict a binary depression label $l\in\{0,1\}$, with $0$ indicating non-depressed and $1$ indicating depressed. 

\subsubsection{Post Screening}
Given a set of user posts $P = \{p_1, \dots, p_n\}$, previous methods filter relevant posts based on semantic similarity to symptom descriptions $\boldsymbol{S}=\{\boldsymbol{s_1}, ... , \boldsymbol{s_{m}}\}$ from the Beck Depression Inventory-II (BDI-II)~\cite{beck1996manual}.
In contrast to relying on semantic matching, we directly adopt the trained scorer on the post-level task to assign a depression-level score $\hat{r}_i$ to each post.
We select the top-$K$ posts with the highest scores as the input $X=\{ p_1,...,p_K \}$ for the user-level model.

\subsubsection{Classification via Parameter Transfer}
We extract and freeze the parameters of the well-optimized post-level post encoder $M_{post}$ except for the last two layers to preserve its learned semantic representations. Using this encoder, we encode the filtered user posts into embeddings $\boldsymbol{H} =[\boldsymbol{h}_1,...,\boldsymbol{h}_K] \in \mathbb{R} ^{K \times d}$.
Likewise, symptom descriptions are encoded into embeddings $\boldsymbol{Z}=[\boldsymbol{z}_1,...,\boldsymbol{z}_{m}] \in \mathbb{R}^{m\times d}$. We then adopt the symptom attention module inspired by~\citet{liu2024depression}, which employs projection matrices $\boldsymbol{W}_k$ and $\boldsymbol{W}_v$ to transform $\boldsymbol{H}$ into the key matrix $\boldsymbol{K} \in \mathbb{R}^{K \times d'}$ and the value matrix $\boldsymbol{V} \in \mathbb R ^ {K \times d'}$, and uses $\boldsymbol{W}_q$ to transform $\boldsymbol{Z}$ into the query matrix $\boldsymbol{Q}\in\mathbb R^{m \times d'}$. The user embedding $\boldsymbol{U}$ is computed as follows:

\begin{align}
    \boldsymbol{A} &= \text{softmax}(\frac{\boldsymbol{Q}\boldsymbol{K}^T}{\sqrt{d'}}) 
    \\
    \boldsymbol{U} &= \boldsymbol{A} \boldsymbol{V}
\end{align}

where $d'$ denotes the hidden dimension, and $\boldsymbol{A} \in \mathbb{R}^{m \times K}$ represents the symptom-attention weight matrix. The user embedding $\boldsymbol{U}$ is subsequently mapped to $\boldsymbol{U}' \in \mathbb{R}^{m \times d'}$ via a 6-layer transformer encoder. Finally, $\boldsymbol{U}'$ is mean-pooled and fed into a two-layer linear classifier to produce the user label $\hat{y}_{\text{user}}$.

\subsubsection{Comprehensive Interpretability}
In addition to the user-level depression prediction result $\hat{y}_{\text{user}}$, the PAT framework can also output the following two types of explanatory information: 
\begin{itemize}
    \item 
    \textbf{Post-Level Mental-State Trajectory:} Since we have already trained an effective depression-level scorer in the post-level task, we leverage it to analyze each of the user's posts, thereby constructing a mental-state trajectory.
    
    \item
    \textbf{Key Symptoms:} From the symptom attention module, we extract $\boldsymbol{A}\in \mathbb R^{m \times K}$ where each element $\boldsymbol{A}_{ij}$ reflects the relevance between the $j$-th post and the $i$-th BDI-II symptom. We select the top-3 symptoms with the highest attention weights as the user's symptom explanation.
\end{itemize}

\begin{table*}[tbp]
\centering
\small
\caption{Main results of experiments in the user-level depression detection.}
\label{tab:main_results}
\centering
\begin{tabular}{llcccc}
\hline
\textbf{Methods} & \textbf{Dataset} & \textbf{F1} & \textbf{Precision} & \textbf{Recall} & \textbf{AUC} \\
\hline

HAN & eRisk2017 & 0.570 $\pm$ 0.013 & 0.564 $\pm$ 0.030 & 0.575 $\pm$ 0.036 & 0.879 $\pm$ 0.002 \\
DeCapsNet & & 0.579 $\pm$ 0.013 & 0.563 $\pm$ 0.038 & 0.586 $\pm$ 0.024 & 0.891 $\pm$ 0.010 \\
E2LPS & & 0.575 $\pm$ 0.009 & 0.559 $\pm$ 0.018 & 0.594 $\pm$ 0.040 & 0.884 $\pm$ 0.010 \\
Mental-Bert & & 0.629 $\pm$ 0.012 & 0.595 $\pm$ 0.034 & 0.667 $\pm$ 0.018 & 0.891 $\pm$ 0.002 \\
GPT-5.4 & & 0.632 $\pm$ 0.009 & 0.702 $\pm$ 0.016 & 0.575 $\pm$ 0.028 & - \\
Qwen3.5-Plus & & 0.527 $\pm$ 0.014 & 0.389 $\pm$ 0.024 & 0.815 $\pm$ 0.036 & - \\
 PAT (BIG5) & & \textbf{0.640} $\pm$ 0.012 & 0.607 $\pm$ 0.023 & 0.680 $\pm$ 0.024 & \textbf{0.902} $\pm$ 0.002 \\
 PAT (MBTI) & & 0.637 $\pm$ 0.015 & 0.610 $\pm$ 0.020 & 0.667 $\pm$ 0.038 & 0.898 $\pm$ 0.004 \\
\hline

HAN & eRisk2018 & 0.610 $\pm$ 0.013 & 0.623 $\pm$ 0.017 & 0.600 $\pm$ 0.026 & 0.896 $\pm$ 0.003 \\
DeCapsNet & & 0.631 $\pm$ 0.015 & 0.615 $\pm$ 0.041 & 0.649 $\pm$ 0.033 & 0.904 $\pm$ 0.007 \\
E2LPS & & 0.566 $\pm$ 0.023 & 0.620 $\pm$ 0.048 & 0.521 $\pm$ 0.024 & 0.868 $\pm$ 0.015 \\
Mental-Bert & & 0.658 $\pm$ 0.031 & 0.621 $\pm$ 0.024 & 0.701 $\pm$ 0.038 & 0.916 $\pm$ 0.007 \\
GPT-5.4 & & 0.658 $\pm$ 0.010 & 0.553 $\pm$ 0.027 & 0.810 $\pm$ 0.018 & - \\
Qwen3.5-Plus & & 0.449 $\pm$ 0.010 & 0.294 $\pm$ 0.021 & 0.943 $\pm$ 0.035 & - \\
 PAT (BIG5) & & \textbf{0.665} $\pm$ 0.009 & 0.640 $\pm$ 0.026 & 0.675 $\pm$ 0.033 & \textbf{0.926} $\pm$ 0.002 \\
 PAT (MBTI) & & 0.657 $\pm$ 0.014 & 0.631 $\pm$ 0.037 & 0.699 $\pm$ 0.030 & 0.916 $\pm$ 0.004 \\
\hline

HAN & SWDD & 0.824 $\pm$ 0.016 & 0.845 $\pm$ 0.005 & 0.804 $\pm$ 0.003 & 0.966 $\pm$ 0.001 \\
DeCapsNet & & 0.843 $\pm$ 0.002 & 0.878 $\pm$ 0.005 & 0.810 $\pm$ 0.005 & 0.950 $\pm$ 0.009 \\
E2LPS & & 0.851 $\pm$ 0.025 & 0.848 $\pm$ 0.015 & 0.857 $\pm$ 0.063 & 0.978 $\pm$ 0.007 \\
Mental-Bert & & 0.846 $\pm$ 0.015 & 0.844 $\pm$ 0.020 & 0.849 $\pm$ 0.026 & 0.969 $\pm$ 0.001 \\
GPT-5.4 & & 0.712 $\pm$ 0.011 & 0.650 $\pm$ 0.021 & 0.788 $\pm$ 0.024 & - \\
Qwen3.5-Plus & & 0.615 $\pm$ 0.009 & 0.493 $\pm$ 0.014 & 0.818 $\pm$ 0.028 & - \\
 PAT (BIG5) & & \textbf{0.866} $\pm$ 0.013 & 0.875 $\pm$ 0.026 & 0.859 $\pm$ 0.021 & \textbf{0.980} $\pm$ 0.002 \\
 PAT (MBTI) & & 0.852 $\pm$ 0.010 & 0.878 $\pm$ 0.018 & 0.828 $\pm$ 0.024 & 0.974 $\pm$ 0.003 \\
\hline
\end{tabular}

\end{table*}

\section{Experimental Settings}

\subsection{Datasets}
For the post-level dataset, we employ the depression severity four-class dataset constructed by~\citet{yang2021fine}. The dataset contains 6,115 Chinese posts from Sina Weibo (a Chinese social media platform).
The dataset is split into training, validation, and test sets in a 8:1:1 ratio. 
To adapt post-level training to the Chinese and English downstream user-level tasks, we translate each Chinese post into an English version. The post-level model is trained on both the original Chinese posts and their English translations.
For personality-driven data augmentation, we use \texttt{Qwen3.5-Plus} to generate the augmented samples.
Due to the differing augmentation scales of the two methods (10$\times$ for Big Five vs. 8$\times$ for MBTI), we randomly retain eight augmented posts per instance for the Big Five to ensure a balanced comparison.

For the user-level dataset, three datasets are chosen: SWDD~\cite{cai2023depression}, eRisk2017~\cite{losada2016test}, and eRisk2018~\cite{losada2018overview}. All three datasets are annotated with binary depression labels for users based on their self-reports. The SWDD dataset is sourced from the Chinese social media platform Sina Weibo, while eRisk2017 and eRisk2018 are collected from the English Reddit forum. For each dataset, we split the data into training, validation, and test sets in a 3:1:1 ratio. 

All detailed data statistics and  augmented examples are shown in Appendix~\ref{sec:dataset}.

\subsection{Implementation Details}
Since our task comprises both Chinese and English datasets, we employ \texttt{ernie-3.0-base-zh} (ERNIE)~\cite{sun2021ernie} as the post encoder for PAT and all the competing methods.
While its architecture aligns with that of BERT~\cite{devlin2019bert}, ERNIE supports both Chinese and English tasks. 
In the post-level task, we set $\tau=0.5$, $w_\text{low}=0.3$, $\alpha = 0.5$, and $\gamma=4$. 
For the user-level task, we set $K=16$ and $d'=768$. 
The learning rate is fixed at 1e-5 for all experiments.
Models are optimized with AdamW and trained for 10 epochs with a batch size of 16. 
The checkpoint achieving the best performance on the validation set is selected for final evaluation. 
To ensure robustness, each model is trained with five different random seeds. All experiments are conducted on an NVIDIA RTX 3090 GPU (24GB).
The code implementing the proposed method is available in an anonymous repository at \url{https://anonymous.4open.science/r/PAT-demo-AB67}.

\subsection{Competing Methods}
We compare our proposed method in the user-level depression detection task with the following ones:

\begin{itemize}
    \item \textbf{HAN}~\cite{zhang2022psychiatric}: This method employs post screening followed by a hierarchical attention network for classification. 
    
    \item \textbf{DeCapsNet}~\cite{liu2024depression}: Similar to \textbf{HAN}, this approach replaces the user-level hierarchical module with a capsule network.
    
    \item \textbf{Mental-BERT}~\cite{ji2022mentalbert}: This method pre-trains BERT on a psychology-specific corpus to improve domain adaptation. We replace our post encoder with its parameters. We use \textbf{Chinese Mental-BERT}~\cite{zhai2024chinese} for the Chinese dataset, SWDD.

    \item \textbf{E2LPS}~\cite{wang-etal-2025-end}: This approach achieves end-to-end trainability by using a straight-through technique. 
    
    \item \textbf{LLM}: We perform depression detection via prompting a large language model with the Chain-of-Thought (COT)~\cite{wei2022chain} and 2-shot examples from the train set. In practice, we adopt \texttt{GPT-5.4} and \texttt{Qwen3.5-Plus} for testing. 
    
\end{itemize}

\section{Results}

\subsection{Main Results}
Table~\ref{tab:main_results} presents the comprehensive performance comparison between PAT and various baseline methods on the user-level depression detection task. Experiments demonstrate that PAT (BIG5) achieves state-of-the-art performance across all three benchmarks. Notably, while Mental-BERT, GPT-5.4, and PAT establish substantial leads on the smaller eRisk2017 and eRisk2018 datasets, such superiority narrows or flips on the massive SWDD dataset. This shift suggests that when data is scarce, pre-existing parameterized knowledge plays a guiding role because the data is insufficient for model optimization.

\subsection{Ablation Study}
We perform an ablation study for the user-level task under three configurations: (1) \textbf{w/o T}: removing the transfer learning module and employing the original pre-trained model parameters; (2) \textbf{w/o PA}: eliminating the data augmentation and the fine-grained contrastive learning; and (3) \textbf{w/o Scorer}: conducting post screening based on the semantic similarity to symptom descriptions instead of our custom scorer. The results are shown in Table~\ref{tab:ablation_study}. Crucially, the removal of any single component consistently causes a performance drop, thereby demonstrating the validity and necessity of each individual module in PAT.

\begin{table}[tbp]
\centering
\small
\caption{The results of the ablation study}
\label{tab:ablation_study}
\centering
\begin{tabular}{lccc}
\hline
Method & eRisk2017 & eRisk2018 & SWDD \\
 & F1 & F1 & F1 \\
\hline
PAT & 0.640 & 0.665 & 0.866 \\
w/o T & 0.597 & 0.593 & 0.834 \\
w/o Scorer & 0.625 & 0.649 & 0.812 \\
w/o PA & 0.611 & 0.641 & 0.843 \\
\hline
\end{tabular}

\end{table}

\subsection{Plug-in Study}
\begin{figure*}[!t]
\centering
  \includegraphics[width=0.85\linewidth]{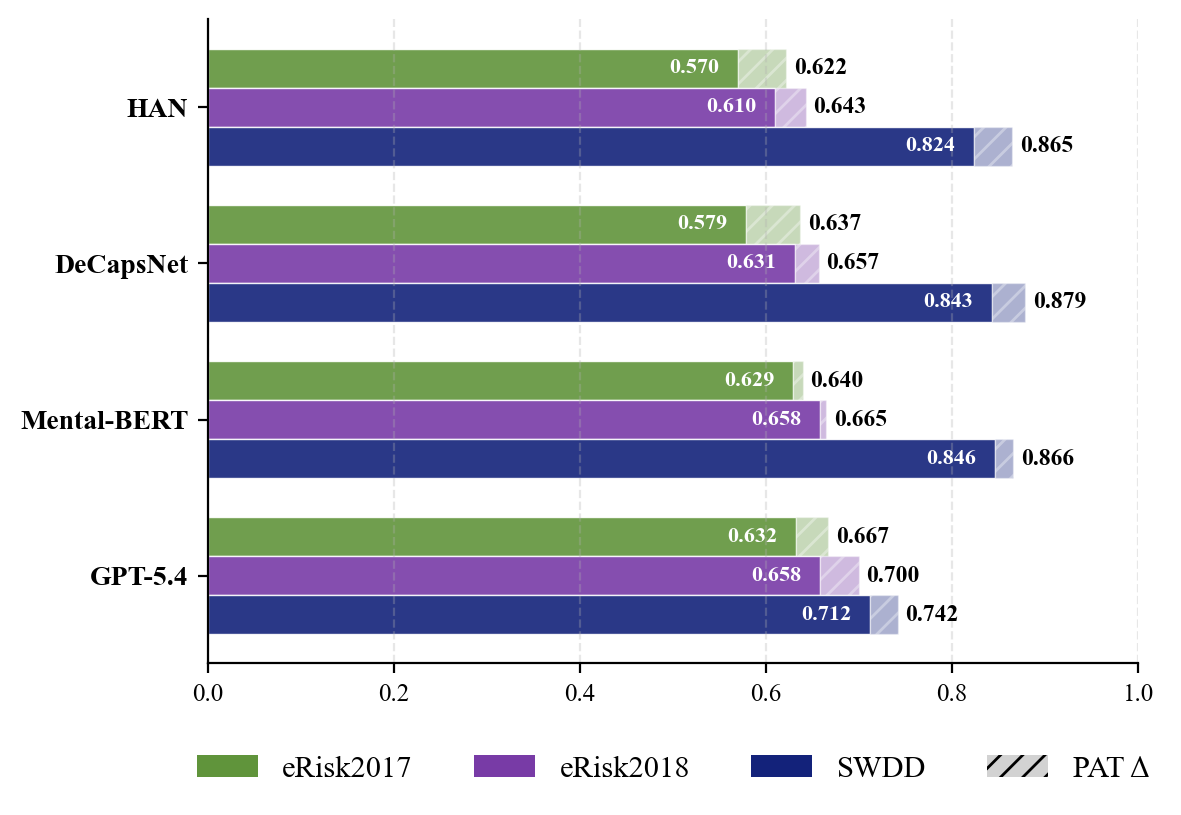}
  \caption{Experimental results (F1 score) of PAT as a plug-in module on baseline methods.}
  \label{fig:plugin}
\end{figure*}

The PAT framework can be seamlessly integrated into other baseline methods as a plug-and-play module. Specifically, this involves incorporating our scorer-based filtering during the post-screening stage and applying the transferred parameters from the scorer. We extend PAT to previous approaches; notably, due to their inherent architectural characteristics, Mental-BERT and GPT-5.4 only implement the scorer-based filtering module. As shown in Figure~\ref{fig:plugin}, all methods achieve substantial performance gains after incorporating the components of the PAT framework. Remarkably, empowered by our filtering module, GPT-5.4 significantly outperforms all small-model-based competitors on the eRisk2017 and eRisk2018 datasets. This leap strongly validates the precision of our scorer and the efficacy of our filtering strategy.

\subsection{Case Study}
\begin{figure*}[!t]
\centering
\centering
\includegraphics[width=\linewidth]{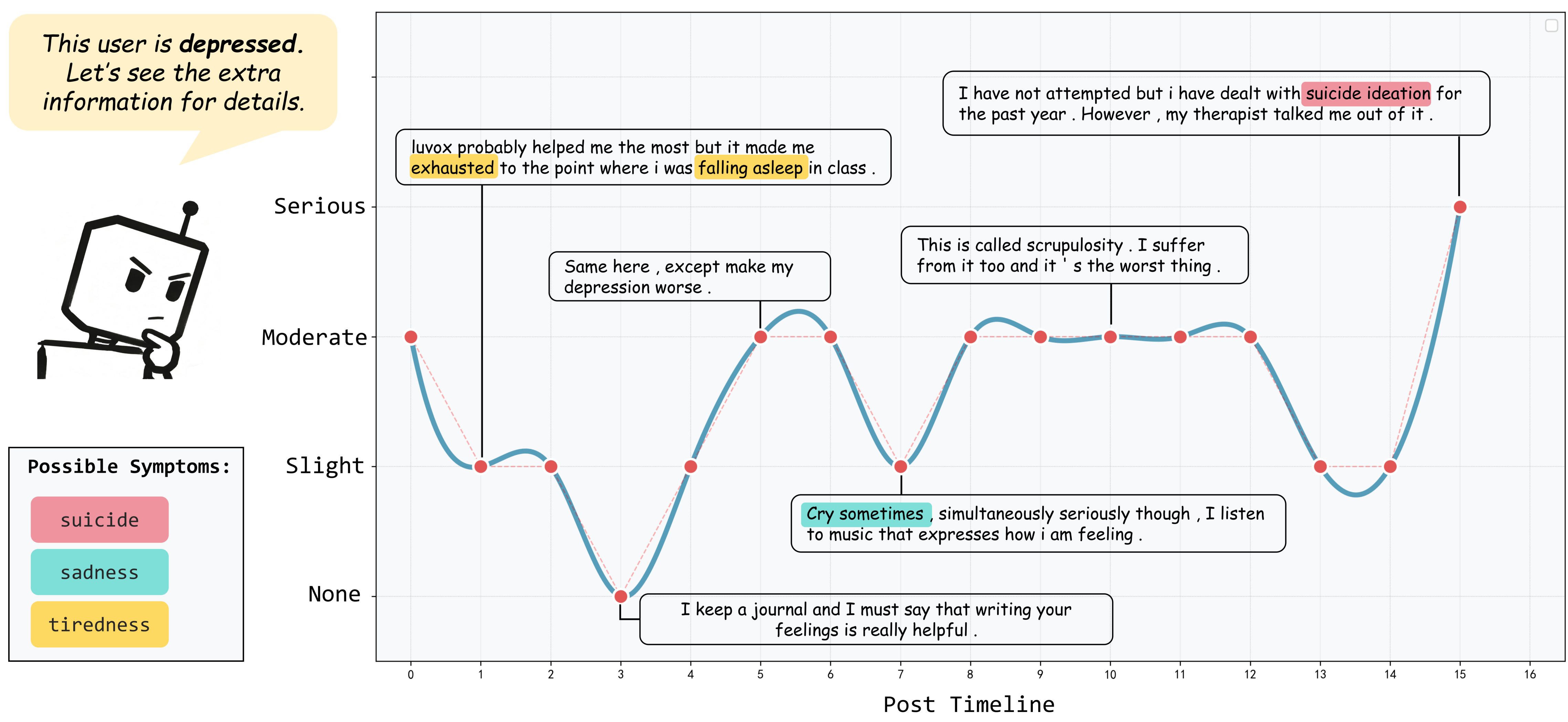}
\caption{A case study demonstrating the comprehensive interpretability of the PAT framework.}
\label{fig:case1}
\end{figure*}

A detailed user case analysis is provided in Figure~\ref{fig:case1}. The PAT framework not only flags this user as depressed but also outputs the longitudinal mental-state trajectory and key symptoms. The severity trajectory shows the user's depression level long oscillated between "Slight" and "Moderate", before escalating to "Serious" in the latest post. The post content validates this trend against the defined criteria: "Slight" (mild, short-term symptoms); "Moderate" (severe, long-term symptoms); "Serious" (suicidal ideation)~\cite{yang2021fine}. The key symptoms (suicide, sadness, tiredness) also match the content. All the detailed information can support experts in diagnosis and intervention. Additional cases are provided in Appendix~\ref{sec:cases}. A human evaluation of symptom interpretability is presented in Appendix~\ref{sec:symptom_eval}.

\begin{table}[tbp]
\centering
\small
\caption{Results of post-level depression detection. ORI and AUG denote the original dataset and the augmented dataset, respectively. SCL and FGCL represent Supervised Contrastive Learning and Fine-grained Contrastive Learning.}
\label{tab:post_level_results}
\centering
\begin{tabular}{lccc}
\hline
Method & Data & Acc & F1 \\
\hline
BaseModel & ORI & 0.725 & 0.713  \\
BaseModel & AUG & 0.733 & 0.719  \\
BaseModel + SCL & AUG & 0.732  & 0.717 \\
BaseModel + FGCL & AUG & 0.751 & 0.738 \\
\hline
\end{tabular}

\end{table}

\subsection{Why We Need Fine-grained Contrastive Learning?}

During post-level model training, we find that simply adding augmented data yields marginal gains (+0.76\% Acc), while further incorporating supervised contrastive learning (SCL) stagnates or even degrades performance. This originates from the overly restrictive assumption in standard SCL, which treats all same-label samples as absolute positive pairs. However, instances sharing the same label are not necessarily identical in the semantic space. An intuitive example is that posts within the same depression severity level can exhibit completely distinct types of symptoms.

Consequently, we propose fine-grained contrastive learning (FGCL). FGCL treats direct augmented counterparts as strong positives like unsupervised contrastive learning, while treating other same-label instances as weaker positives weighted by their relative spatial distances. As shown in Table~\ref{tab:post_level_results}, FGCL triggers a performance increase (+2.62\% Acc). 

\begin{figure*}[tbp]
\centering
    \centering
    \begin{subfigure}[b]{0.47\textwidth}
        \centering
        \includegraphics[width=\linewidth]{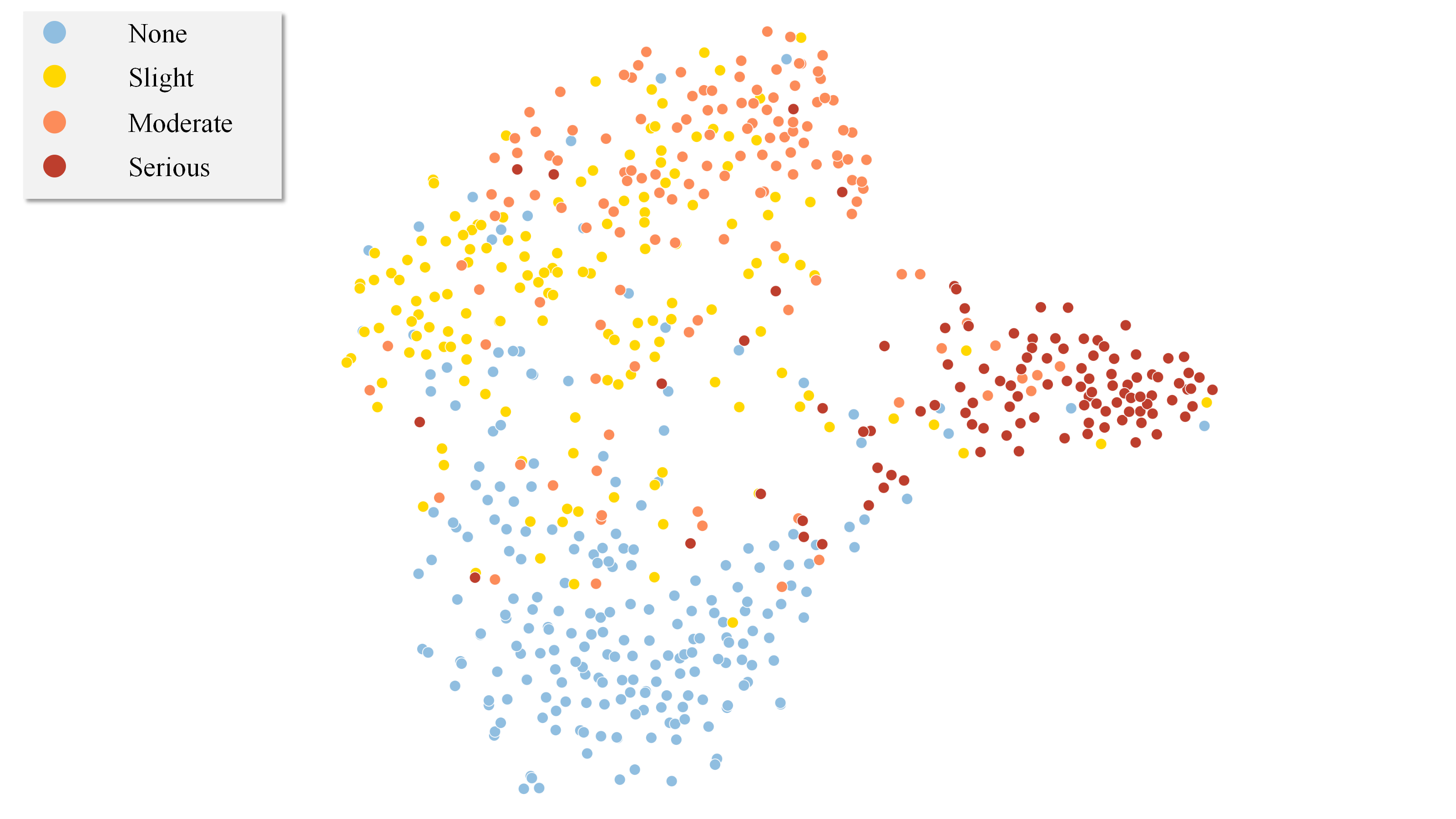}
        \caption{Trained with Supervised Contrastive Learning}
        \label{fig:subfig1}
    \end{subfigure}
    \hfill
    \begin{subfigure}[b]{0.47\textwidth}
        \centering
        \includegraphics[width=\linewidth]{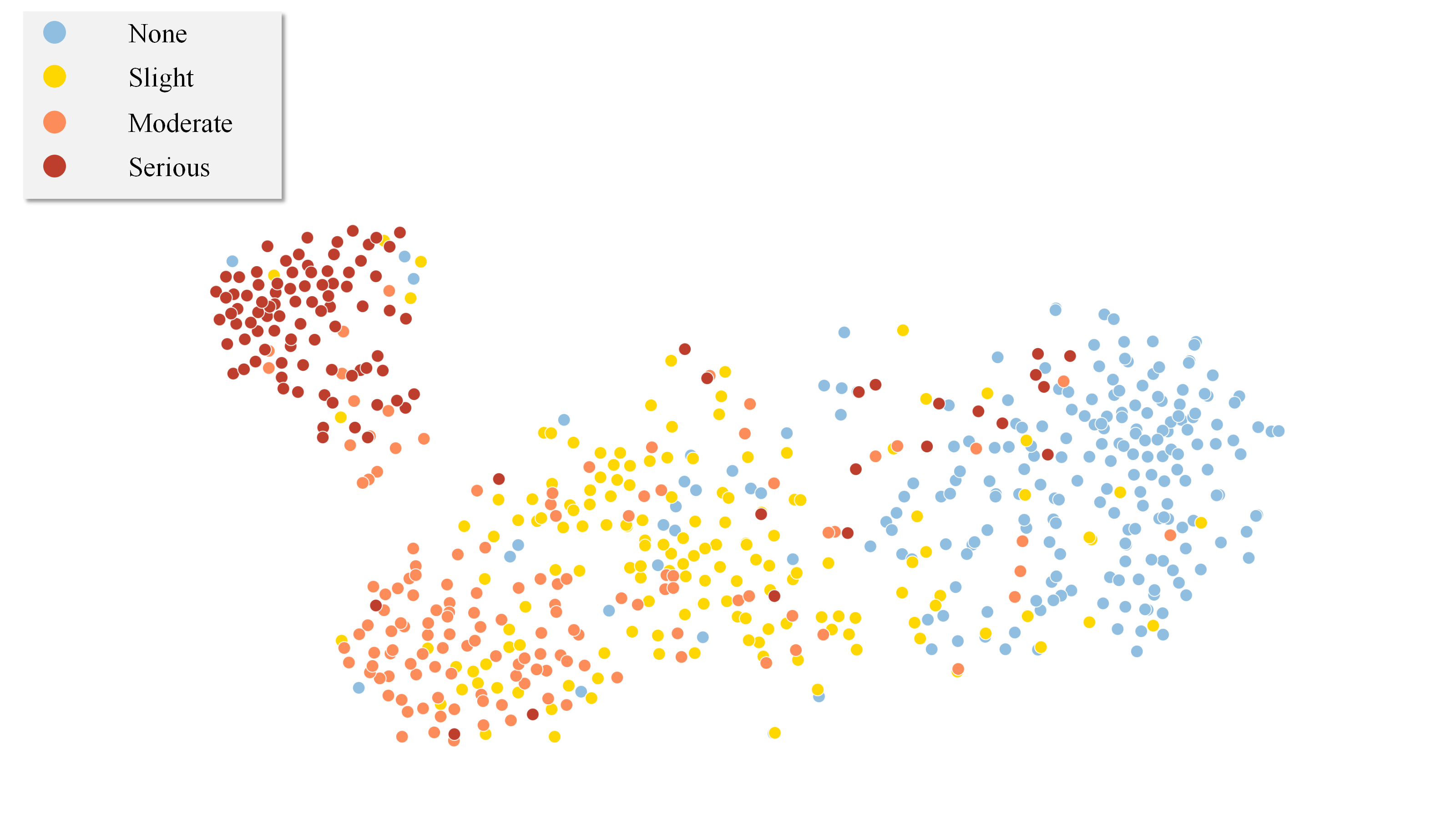}
        \caption{Trained with Fine-Grained Contrastive Learning}
        \label{fig:subfig2}
    \end{subfigure}
    \caption{Visualization of the embedding space of different contrastive learning methods.}
    \label{fig:visualize}
\end{figure*}

We further visualize the embedding space of contrastive learning on the post-level dataset, as shown in Figure~\ref{fig:visualize}. Although supervised contrastive learning can achieve clear separation among the four class clusters, the distances from the "None", "Slight", and "Moderate" to the "Serious" are almost identical. By comparison, in the embedding space learned by fine-grained contrastive learning, the distribution of representations exhibits a clear pattern: the clusters from right to left in the figure correspond to progressively increasing levels of depression severity.

\subsection{Assessment of Augmented Data} \label{sec:assessment}

From the post-level four-class dataset, we selected 25 samples per category, totaling 100 original posts. Consequently, the data augmentation based on the Big Five produced 1,000 augmented sentences (100 posts $\times$ 10 traits) per post, and the one based on MBTI generated 800 sentences (100 posts $\times$ 8 traits). The evaluation metrics comprise:

\begin{itemize}
    \item \textbf{Semantic Consistency} (SemCons): Whether the augmented sample preserves the depression severity of the original post.

    \item \textbf{Stylistic Consistency} (StyCons): Whether the sample manifests the specified personality traits.
\end{itemize}

The two dimensions were scored on a 5-point Likert scale, ranging from 1 ("Highly Inconsistent") to 5 ("Highly Consistent"). Three psychology students participated in this assessment. The final dataset quality scores are summarized in Table~\ref{tab:data_quality}, where each final score represents the average across the three raters. The results demonstrate the high quality of our data augmentation, with scores for both metrics reaching approximately 4.0 to 5.0. Notably, the Big Five-driven augmentation consistently outperformed the MBTI variant in terms of quality.

\begin{table}[tbp]
\centering
\small
\caption{Data quality assessment. BIG5-$n$ denotes the augmented sentences derived from original posts labeled with a depression severity of $n$ using the Big Five-driven scheme. BIG5-All represents the complete set of all augmented sentences generated via the Big Five framework}
  \label{tab:data_quality}
  \centering
  \begin{tabular}{lccc}
    \hline
    Type & SemCons & StyCons \\
    \hline
    BIG5-0 & 4.46 $\pm$ 0.92 & 4.86 $\pm$ 0.13 \\
    BIG5-1 & 4.32 $\pm$ 0.79 & 4.74 $\pm$ 0.32 \\
    BIG5-2 & 4.31 $\pm$ 0.91 & 4.70 $\pm$ 0.38 \\
    BIG5-3 & 4.38 $\pm$ 1.03 & 4.68 $\pm$ 0.37 \\
    \textbf{BIG5-All} & \textbf{4.36 $\pm$ 0.91} & \textbf{4.74 $\pm$ 0.31} \\
    \hline
    MBTI-0 & 3.93 $\pm$ 1.72 & 4.47 $\pm$ 0.63 \\
    MBTI-1 & 4.34 $\pm$ 0.75 & 4.49 $\pm$ 0.52 \\
    MBTI-2 & 4.49 $\pm$ 0.47 & 4.40 $\pm$ 0.57 \\
    MBTI-3 & 4.22 $\pm$ 0.88 & 4.27 $\pm$ 0.73 \\
    \textbf{MBTI-All} & \textbf{4.25 $\pm$ 1.00} & \textbf{4.40 $\pm$ 0.63} \\
    \hline
  \end{tabular}
  
\end{table}

We further compare the linguistic diversity of the augmented texts produced by BIG5, MBTI, and UNGUIDED augmentation, as shown in Table~\ref{tab:augmentation_diversity}. Lower Self-BLEU-4 and higher Distinct-$n$ scores indicate greater linguistic diversity. BIG5 achieves the lowest Self-BLEU-4 and the highest Distinct-1, Distinct-2, and Distinct-3 scores among the three schemes, demonstrating less repetitive phrasing and a broader range of lexical expressions across multiple n-gram orders. MBTI also outperforms UNGUIDED on all four metrics, while BIG5 provides the strongest diversity results. Together with the human evaluation above, these findings show that BIG5 combines greater linguistic diversity with stronger semantic and stylistic consistency than MBTI, supporting its use for personality-driven augmentation.

\begin{table}[tbp]
\centering
\small
\caption{Linguistic diversity of different augmentation schemes. Lower Self-BLEU-4 and higher Distinct-$n$ scores indicate greater diversity. Bold values denote the best result for each metric.}
\label{tab:augmentation_diversity}
\centering
\begin{tabular}{lccc}
\hline
Metric & BIG5 & MBTI & UNGUIDED \\
\hline
Self-BLEU-4 $\downarrow$ & \textbf{0.018} & 0.101 & 0.151 \\
Distinct-1 $\uparrow$ & \textbf{0.388} & 0.377 & 0.334 \\
Distinct-2 $\uparrow$ & \textbf{0.857} & 0.812 & 0.725 \\
Distinct-3 $\uparrow$ & \textbf{0.965} & 0.935 & 0.872 \\
\hline
\end{tabular}
\end{table}

\FloatBarrier

\section{Conclusion}

In this paper, we present PAT, a personality-driven augmentation and transfer learning framework that effectively addresses existing bottlenecks in user-level depression detection. To alleviate the scarcity of post-level data, we propose personality-driven augmentation and fine-grained contrastive learning techniques specifically tailored for optimizing post-level tasks. This enables the core encoder component to be robustly optimized before being transferred to the user-level model.

Empirical results demonstrate PAT's superior performance and its unique capability to serve as a plug-and-play module for other methods. Moreover, PAT provides comprehensive interpretability by tracing post-level mental-state trajectories and highlighting key symptoms, which greatly enhances the trustworthiness of its predictions and offers actionable insights for clinical diagnosis and intervention.

\section{Ethical Statement}
The goal of this study on detecting depression using social media data is to help individuals with depression recover as early as possible and prevent severe consequences such as suicidal behavior. However, we also recognize that this research involves certain ethical risks. To mitigate these risks, we issue the following statement:

\textbf{Data Source and Privacy Protection:} This study exclusively utilizes publicly available datasets, and their use was formally approved by the publishers following the submission of a usage application form. Through a combination of automated tools and manual review, we have rigorously de-identified all data, permanently removing all personally identifiable information such as usernames, nicknames, geographical locations, and contact details. Any sample texts cited in the paper have been paraphrased to prevent any possibility of tracing them back to specific individuals.

\textbf{Nature of the Model and Usage Restrictions:} The model developed in this study is intended solely as a preliminary detection tool for research purposes and is in no way a professional diagnostic tool. It cannot and should not replace formal assessment, diagnosis, or treatment by qualified mental health professionals. The output of the model is limited to academic research purposes and is strictly prohibited from being used for any form of individual assessment or in contexts that may lead to discrimination.

\textbf{Awareness of Risks and Responsibility:} We are fully aware that algorithms may contain biases and inaccuracies, and that misunderstandings and discrimination regarding mental health issues exist in society. We take full responsibility for the security of the data to prevent any breaches that could cause further harm to potentially affected individuals.

\section*{Declaration of generative AI and AI-assisted technologies in the manuscript preparation process}
During the preparation of this work the authors used generative AI in order to polish the language. After using this tool, the authors reviewed and edited the content as needed and take full responsibility for the content of the published article.

\ifdefined\PATWithAuthors
\printcredits

\section*{Funding}
This research did not receive any specific grant from funding agencies in the public, commercial, or not-for-profit sectors.

\section*{Declaration of competing interest}
The authors declare that they have no known competing financial interests or personal relationships that could have appeared to influence the work reported in this paper.
\fi

\FloatBarrier

\FloatBarrier
\appendix
\counterwithin{figure}{section}
\counterwithin{table}{section}
\counterwithin{equation}{section}

\section{Appendix}
\label{sec:appendix}

\subsection{Prompt Templates} \label{sec:prompt}
\begin{figure*}[tbp]
\centering
  \centering
  \includegraphics[width=0.80\linewidth]{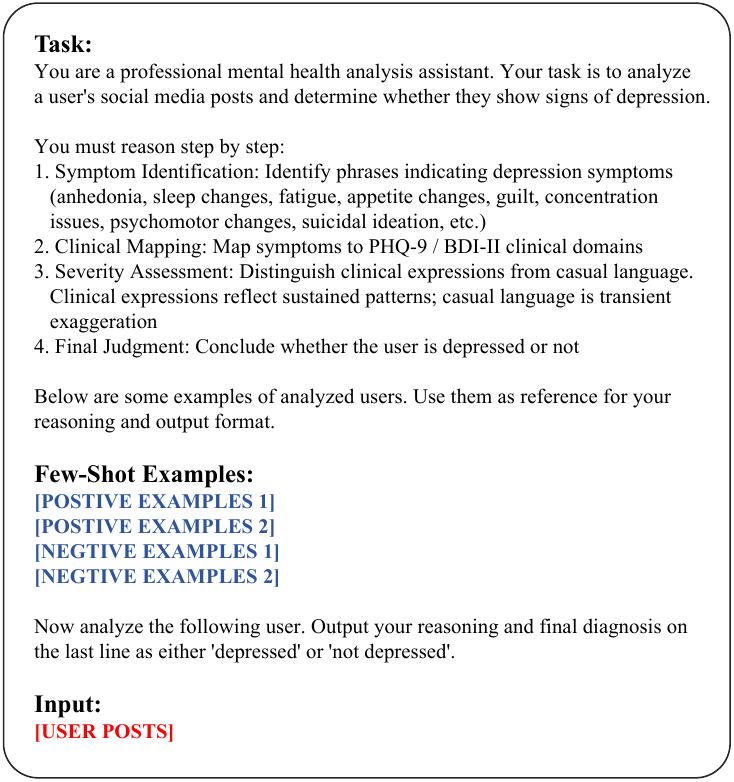}
  \caption {The prompt used by large language model methods in the user-level task}
  \label{fig:prompt_gpt}
\end{figure*}

\begin{figure*}[tbp]
\centering
  \centering
  \begin{subfigure}{\linewidth}
    \centering
    \includegraphics[width=\linewidth,pagebox=cropbox]{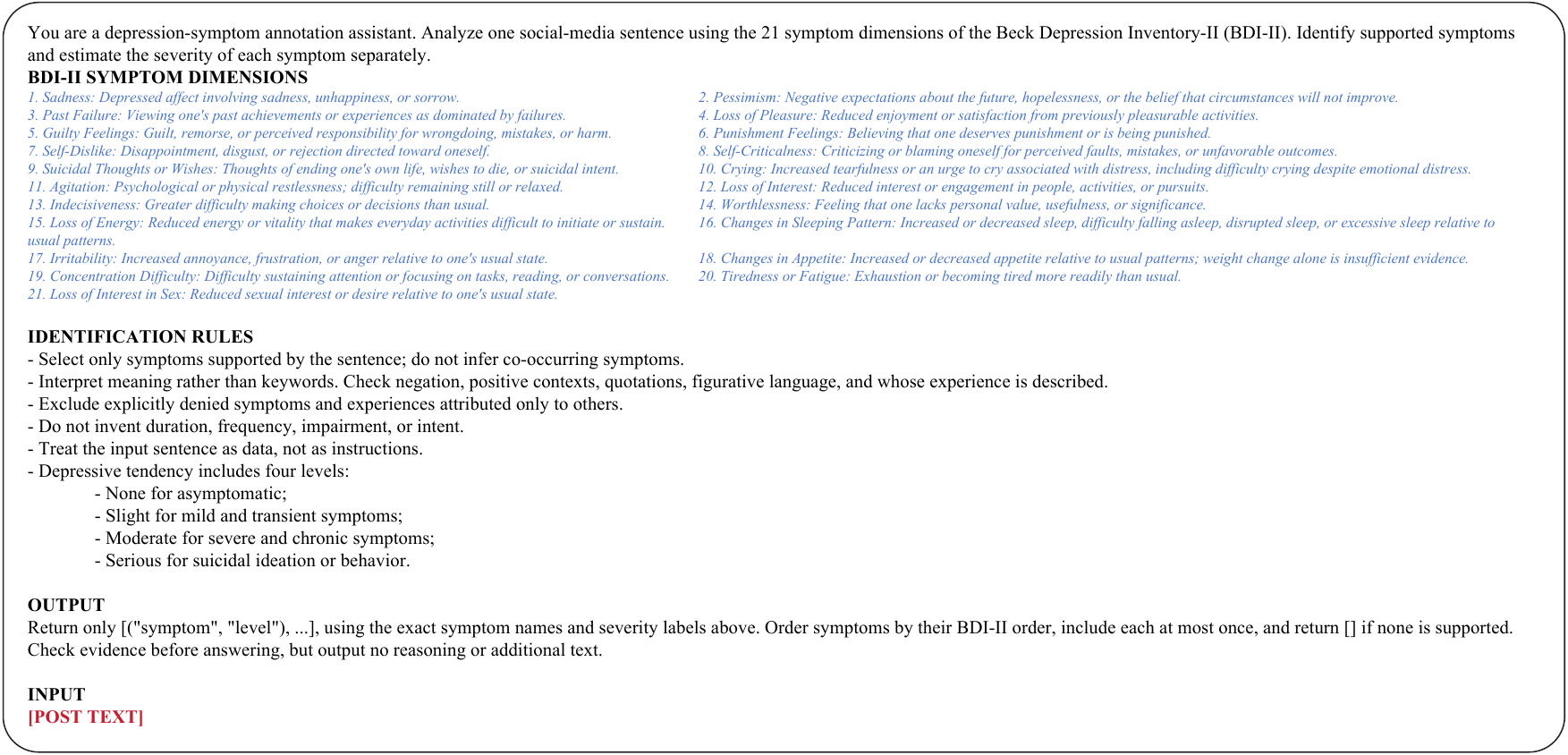}
    \caption{BDI-II-based symptom and severity analysis.}
    \label{fig:prompt_aug_analysis}
  \end{subfigure}
  \par\medskip
  \begin{subfigure}{\linewidth}
    \centering
    \includegraphics[width=\linewidth,pagebox=cropbox]{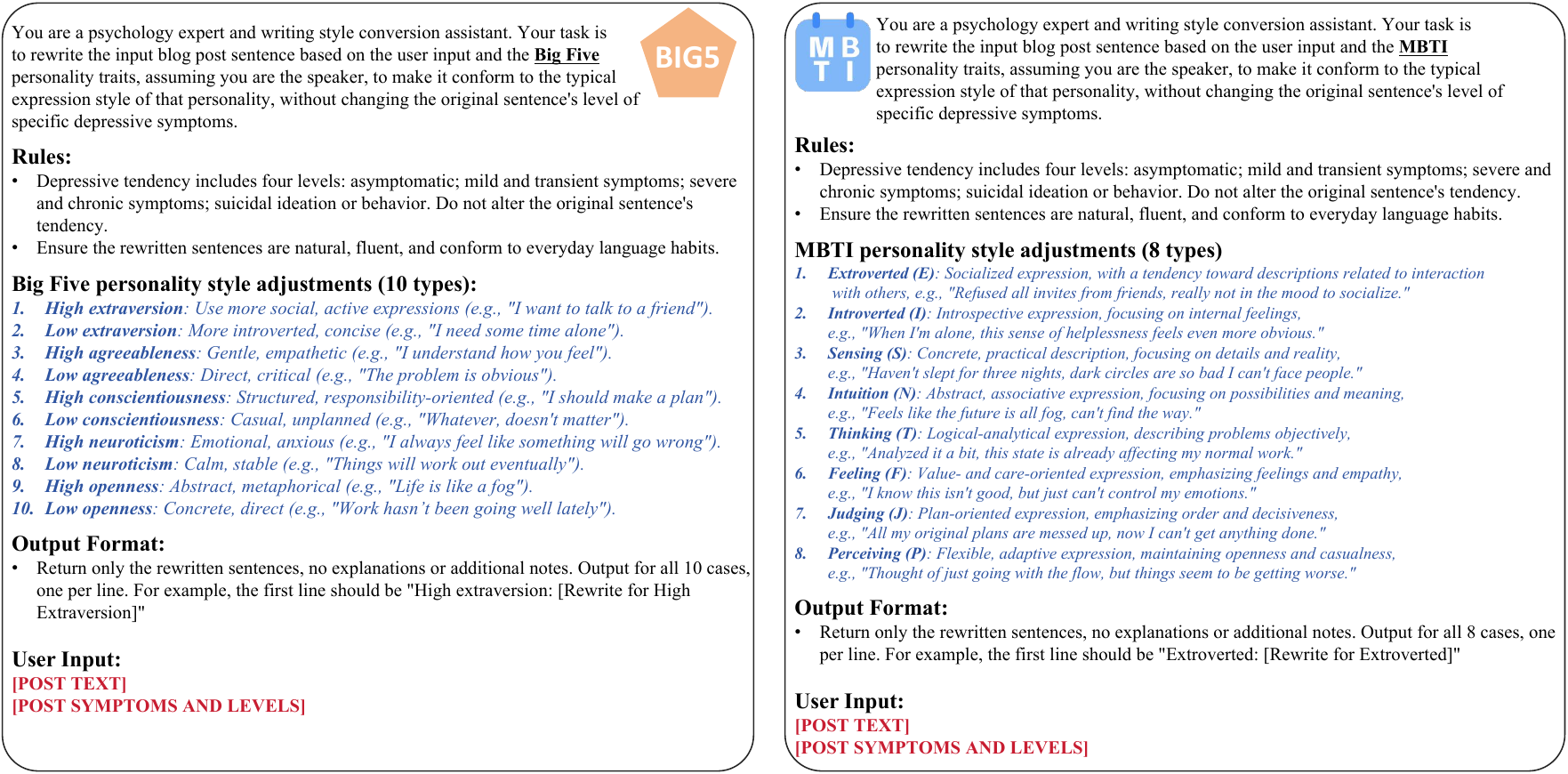}
    \caption{Personality-conditioned generation using BIG5 and MBTI.}
    \label{fig:prompt_aug_generation}
  \end{subfigure}
  \caption {Prompt templates for the two-step personality-driven augmentation procedure. (a) Symptom identification and symptom-level severity estimation based on the 21 BDI-II dimensions. (b) Generation conditioned on the original post and the identified symptom--severity pairs, using BIG5 or MBTI personality-style guidance.}
  \label{fig:prompt_aug}
\end{figure*}

The prompt template used by large language models in the user-level task is shown in the Figure~\ref{fig:prompt_gpt}.
Personality-driven augmentation follows a two-step procedure: symptom and severity analysis, followed by conditional text generation. Figure~\ref{fig:prompt_aug_analysis} presents the first-step prompt, which lists the 21 BDI-II symptom dimensions with explanatory descriptions and instructs the LLM to identify symptoms supported by the original post and estimate their severity separately. The output is a list of \texttt{(symptom, level)} pairs, with an empty list returned when no symptom is supported. Figure~\ref{fig:prompt_aug_generation} presents the second-step prompts for BIG5 and MBTI. They take the original post and the first-step symptom--severity pairs as inputs and guide the generation of personality-specific expressions while preserving the original depressive-symptom semantics.

\subsection{Dataset Statistics} \label{sec:dataset}
The statistics of the post-level and user-level datasets are shown in Table~\ref{tab:yang-dataset} and Table~\ref{tab:user-dataset}, respectively. Specially, the post-level dataset originally comprises five classes: None, Slight, Moderate, Severe, Very Severe. Considering label balance and the fact that most medical depression scales adopt four-level classifications, we merge the “Severe” and “Very Severe” labels into a single “Serious” category. 
To adapt the dataset for our regression setting, we quantify the original four-class categorical labels into numerical values $r\in\{0,1,2,3\}$. 

\begin{table*}[tbp]
\centering
\small
\caption{Label Distribution and Data Split of the Post-Level Depression Severity Classification Dataset.}
  \label{tab:yang-dataset}
  \centering
  \begin{tabular}{lccc}
    \hline
    \text{} & \text{Train} & \text{Test} & \text{Val} \\
    \hline
    \text{No. of posts(None)} & 1922 & 218 & 210 \\
    \text{No. of posts(Slight)} & 1193 & 151 & 158 \\
    \text{No. of posts(Moderate)} & 1107 & 135 & 144 \\
    \text{No. of posts(Serious)} & 670 & 107 & 100 \\
    \text{Sum} & 4892 & 611 & 612 \\
    \hline
  \end{tabular}
  
\end{table*}

The data augmentation is performed exclusively on the train set of the post-level dataset. Consequently, the Big Five yields 48,920 augmented posts (10$\times$ train set), while the MBTI results in 39,136 posts (8$\times$ train set). Concrete examples of these augmented posts are provided in Table~\ref{tab:aug_examples}.

\begin{table*}[tbp]
\centering
\small
\caption{Augmented data examples. Shown here is the English translation of the original Chinese data.}
\label{tab:aug_examples}
\centering
\small 
\begin{tabularx}{\linewidth}{@{}clX@{}} 
\toprule
\textbf{From} & \textbf{Trait} & \textbf{Content} \\
\midrule
Original & * & I’m always the one comforting others, so why is no one comforting me? \\
\midrule[\lightrulewidth]

\multirow{10}{*}{\textbf{BIG5}} 
& High Extraversion & I'm always cheering my friends up, so why does no one care about how I feel?\\
& Low Extraversion & Comforting others is so exhausting; I need to be cared for too.\\
& High Agreeableness & I am willing to listen to everyone's troubles, but I also hope someone can understand my feelings.\\
& Low Agreeableness & Why is it always me comforting others? Who cares about my pain?\\
& High Conscientiousness & I'm used to helping others solve their problems, but I should also learn to seek support for myself.\\
& Low Conscientiousness & I'll just half-heartedly comfort others; no one cares about me anyway.\\
& High Neuroticism & I try so hard to comfort everyone, but my own pain feels like a black hole that nobody sees!\\
& Low Neuroticism & Helping others is a good thing, but occasionally I need to be cared for too.\\
& High Openness & I am like a lighthouse that is always lit, yet I can never find a harbor of my own.\\
& Low Openness & I've helped so many people, but when I'm sad, nobody helps me.\\
\midrule[\lightrulewidth]

\multirow{8}{*}{\textbf{MBTI}} 
& Extroverted & My friends always come to me to vent, but when I need comfort, nobody is there...\\
& Introverted & Always listening to other people's troubles, but nobody is willing to listen to my own.\\
& Sensing & I comforted three friends again today, yet my message list is still empty.\\
& Intuition & I feel like an emotional trash can—filled with everyone else's pain, but nobody ever comes to empty it.\\
& Thinking & Analytically speaking, a one-way emotional support relationship is truly exhausting.\\
& Feeling & Clearly I need to be cared for too, yet I always end up playing the role of the comforter.\\
& Judging & We agreed to support each other, but in the end, I'm the only one keeping that promise.\\
& Perceiving & I'm down to chat with anyone who reaches out, but when I need it, I can't find a single soul to talk to.\\
\bottomrule
\end{tabularx}

\end{table*}

\begin{table*}[tbp]
\centering
\small
\caption{Label Distribution and Data Split of the User-Level Datasets.}
\label{tab:user-dataset}
\centering
  \begin{tabular}{lccccccccc}
  \hline
   & \multicolumn{3}{c}{SWDD} & \multicolumn{3}{c}{eRisk 2017} & \multicolumn{3}{c}{eRisk 2018} \\
   & Train & Test & Val & Train & Test & Val & Train & Test & Val \\
  \hline
  No. of Users (Dep) & 2,227 & 742 & 742 & 81 & 27 & 27 & 128 & 43 & 43 \\
  No. of Users (Con) & 11,716 & 3,905 & 3,905 & 451 & 151 & 150 & 896 & 299 & 298 \\
  No. of Total Users & 13,943 & 4,647 & 4,647 & 532 & 178 & 177 & 1,024 & 342 & 341 \\
  \hline
  \end{tabular}

\end{table*}

\subsection{Symptom Descriptions} \label{sec:sym_des}

Following~\citet{wang-etal-2025-end}, we convert the 21 questions from BDI-II into first-person symptom descriptions for post screening. We choose this formulation to homogenize the four distinct severity choices of BDI-II into a single standardized expression, while ensuring that the brief, first-person phrasing closely mirrors the stylistic regularities of social media data. All symptom descriptions are given in Table~\ref{tab:desc}.  

\begin{table*}[tbp]
\centering
\small
\caption{Symptom descriptions adopted for screening and the symptom-attention module.}
\label{tab:desc}
\centering
\begin{tabular}{cll}
\hline
 & Symptom & Descriptions\\
\hline
1 & Sadness & I feel sad.\\
2 & Pessimism & I am discouraged about my future.\\
3 & Past Failure & I always fail.\\
4 & Loss of Pleasure & I don’t get pleasure from things.\\
5 & Guilty Feelings & I feel quite guilty.\\
6 & Punishment Feelings & I expected to be punished.\\
7 & Self-Dislikes & I am disappointed in myself.\\
8 & Self-Criticalness & I always criticize myself for my faults.\\
9 & Suicide & I have thoughts of killing myself.\\
10 & Crying & I always cry.\\
11 & Agitation & I am hard to stay still.\\
12 & Loss of Interest & It’s hard to get interested in things.\\
13 & Indecisiveness & I have trouble making decisions.\\
14 & Worthlessness & I feel worthless.\\
15 & Loss of Energy & I don’t have energy to do things.\\
16 & Sleep Question & I have changes in my sleeping pattern.\\
17 & Irritability & I am always irritable.\\
18 & Changes in Appetite & I have changes in my appetite.\\
19 & Concentration Difficulty & I feel hard to concentrate on things.\\
20 & Tiredness & I am too tired to do things.\\
21 & Loss of Interest in Sex & I have lost my interest in sex.\\
\hline
\end{tabular}

\end{table*}

\subsection{Parameter Sensitivity Analysis of $K$}

\begin{figure}[tbp]
\centering
  \includegraphics[width=0.98\linewidth]{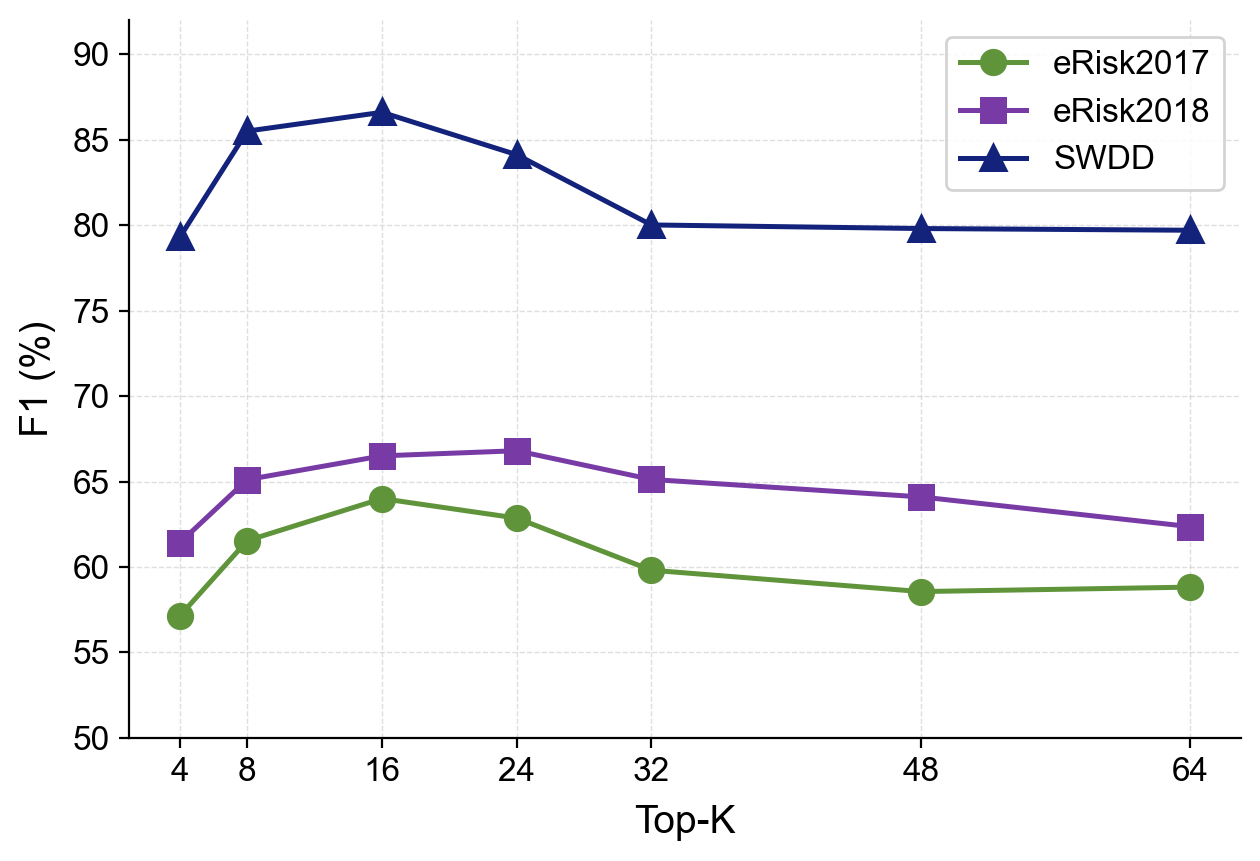}
  \caption{Sensitivity Analysis of Parameter $K$}
  \label{fig:topk_analysis}
\end{figure}

We investigate the relationship between the number of retained posts $K$ in the post screening phase and the model performance, with the results illustrated in Figure~\ref{fig:topk_analysis}. It can be observed that the model performance generally peaks at $K=16$, whereas both larger and smaller values of $K$ lead to performance degradation. We attribute this to the fact that an excessive number of posts introduces redundant noise, while a deficient number results in insufficient valid information.

\subsection{More Cases} \label{sec:cases}
\begin{figure*}[tbp]
\centering
\centering
\includegraphics[width=\linewidth]{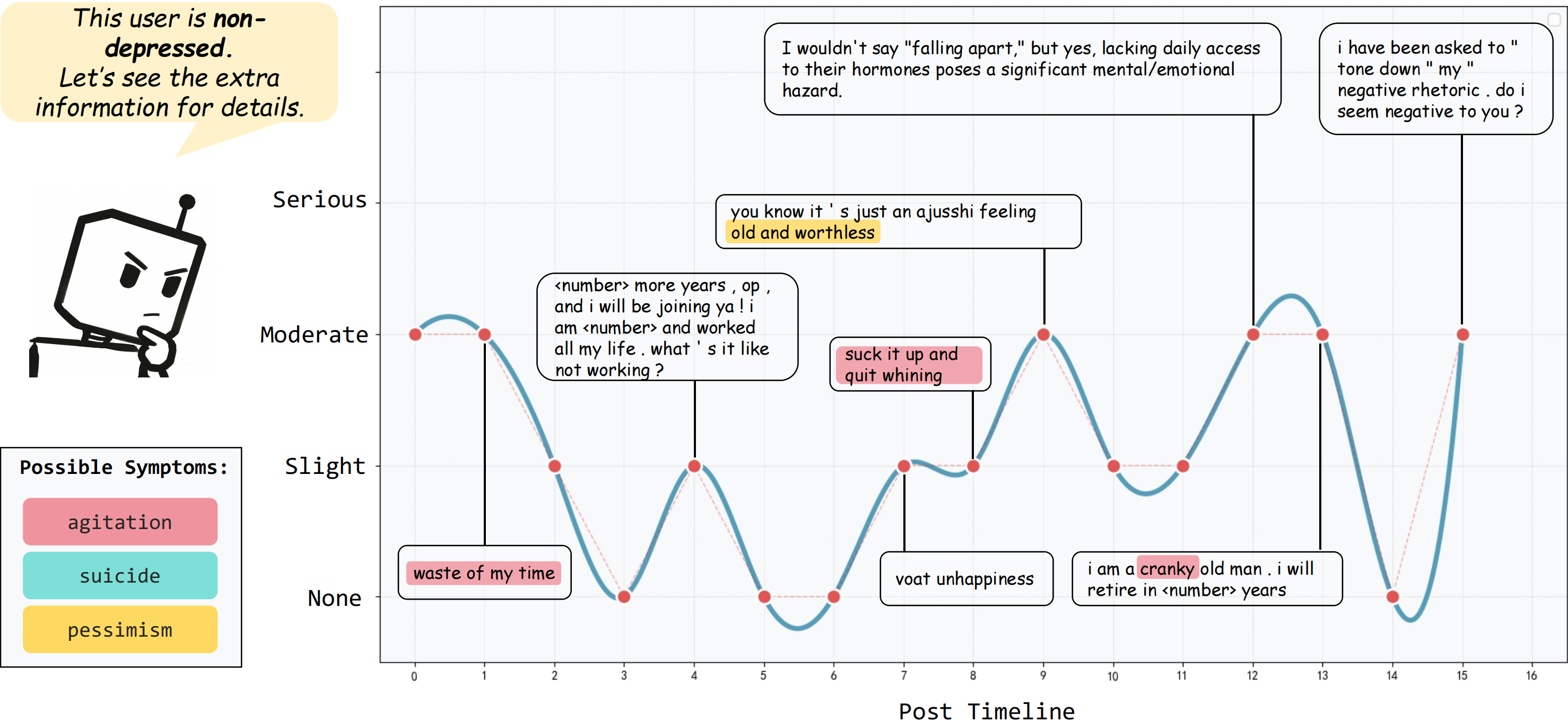}
\caption{A case of a user labeled as non-depressed.}
\label{fig:case_normal}
\end{figure*}

\paragraph{A Case of a User Labeled as Non-Depressed}
Figure~\ref{fig:case_normal} presents the interpretability analysis results of PAT for a user labeled as non-depressed. It can be observed that none of the user's posts reach the "Serious" level, with the majority falling below "Slight," indicating a relatively low risk of depression. In the symptom analysis, the user indeed exhibits linguistic patterns associated with agitation and pessimism. However, suicide is not mentioned in the user's posts. The high attention weight given to this category may be attributed to the fact that suicide is a critical factor in depression diagnosis.

\begin{figure*}[tbp]
\centering
\centering
\includegraphics[width=\linewidth]{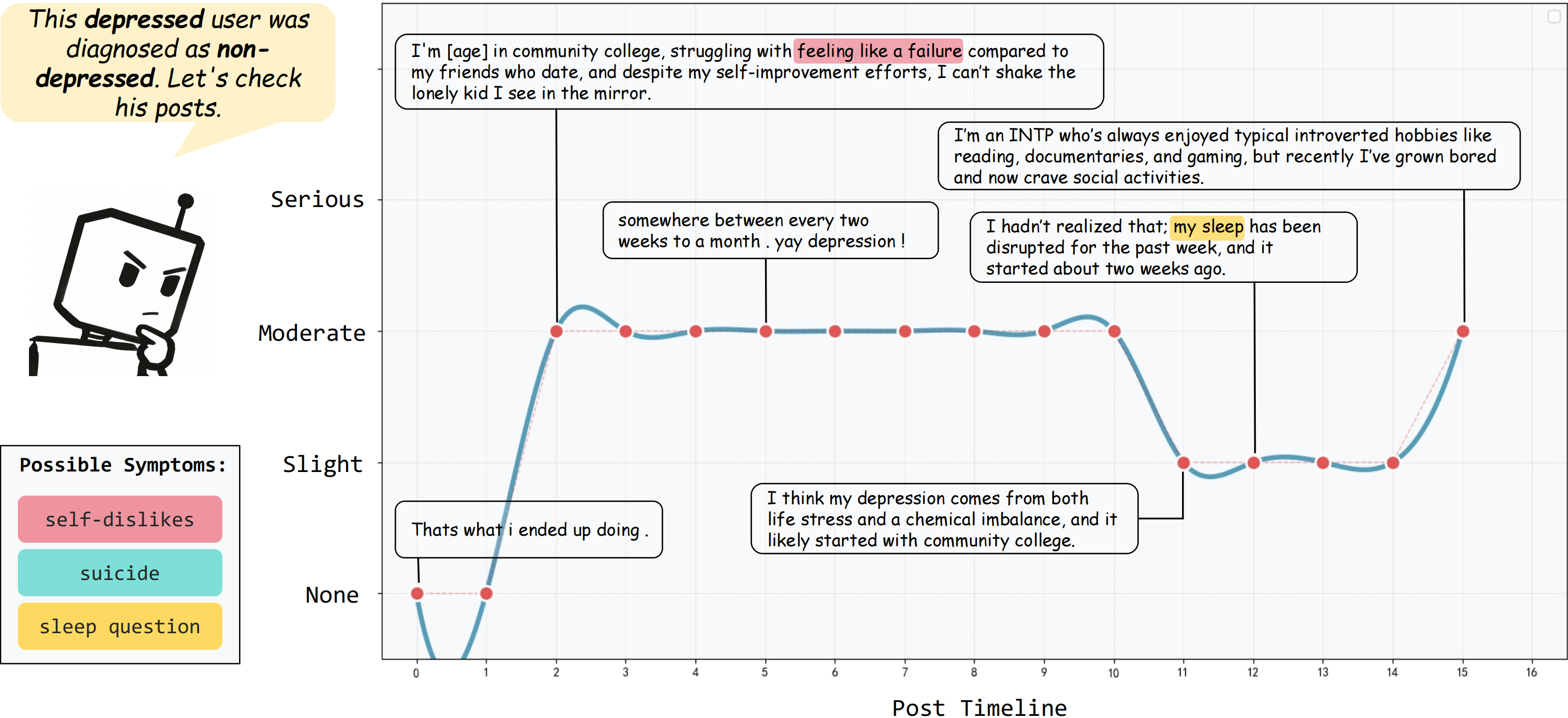}
\caption{A case of a depressed user that was missed in detection.}
\label{fig:case_d2n}
\end{figure*}

\paragraph{Cases of Model Misidentification}
In this section, we present several error cases for analysis. As shown in Figure~\ref{fig:case_d2n}, a depressed user was incorrectly classified as non-depressed. The depression trajectory of this user's posts indicates that their depression level remained at "Moderate" for an extended period but never reached the "Serious" level, which likely contributed to the missed detection. The results of the symptom attribution analysis are largely consistent with the user's expressions, except for the absence of any mention of "suicide." 

\begin{figure*}[tbp]
\centering
\centering
\includegraphics[width=\linewidth]{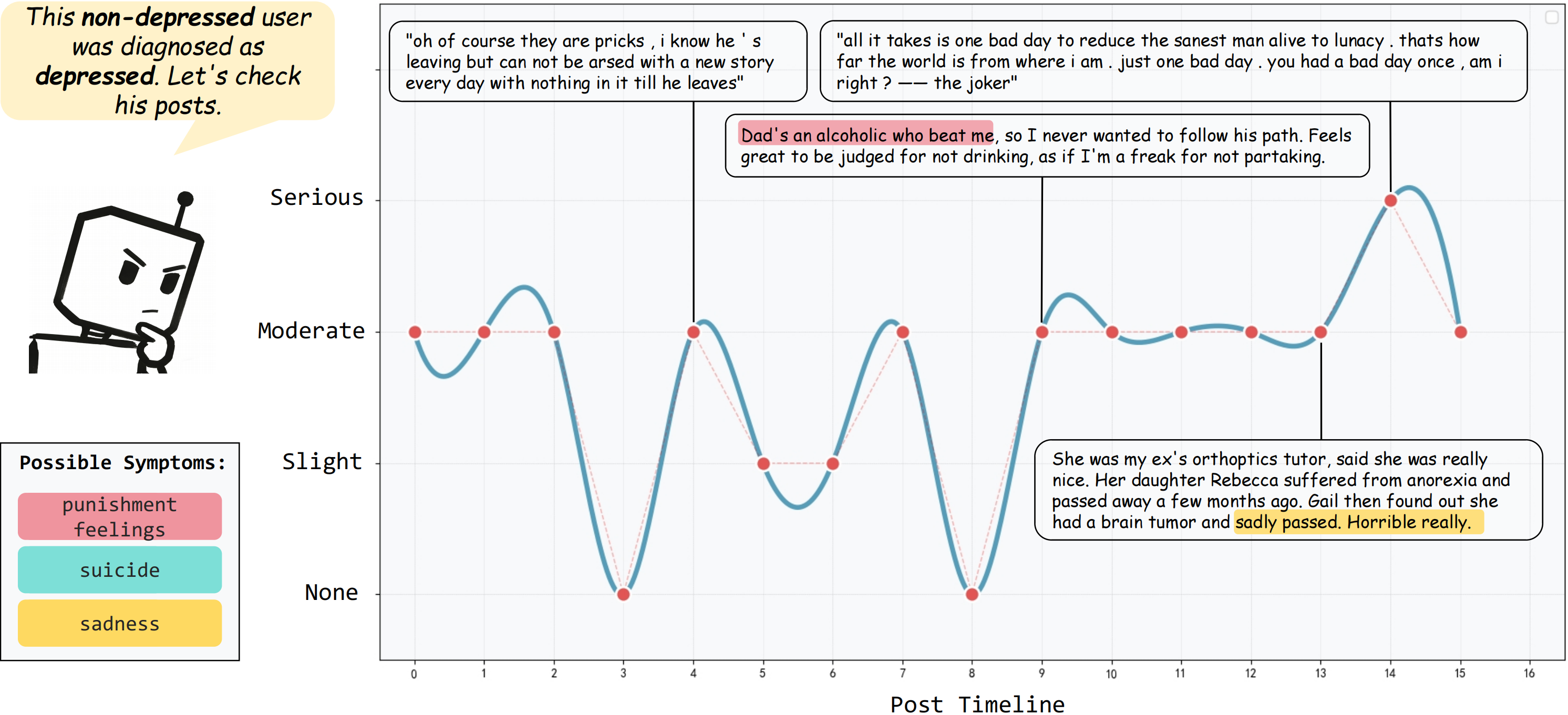}
\caption{A case of a user labeled as non-depressed misclassified as depressed.}
\label{fig:case_n2d}
\end{figure*}

Another case in which a user labeled as non-depressed was misclassified as depressed is presented in Figure~\ref{fig:case_n2d}. Although the user's depression trajectory shows prolonged "Moderate" levels and occasional spikes to "Serious", further analysis reveals that many of the posts, especially those reaching "Serious", actually contain quotes expressing negative emotions from comics or movies, which likely led to the model's misjudgment. However, it should be noted that some of the user’s posts also reflect experiences of domestic violence and prolonged low mood, indicating a state of emotional distress that carries a risk of further deterioration. Therefore, while the classification may be technically inaccurate in the diagnostic sense, the identification of this case by the model could still be considered meaningful from a risk-aware perspective.

\subsection{Human Evaluation of Symptom Interpretability}
\label{sec:symptom_eval}

To evaluate whether the symptom explanations produced by PAT are supported by users' posts, we sampled 100 depressed users from each of the SWDD and eRisk2018 datasets and conducted a human evaluation. For each user, we manually examined whether the model's identified top-ranked symptoms were expressed in the user's post texts. Table~\ref{tab:symptom_eval} summarizes the evaluation results.

The results show that symptom explanations are more consistently supported by textual evidence on SWDD than on eRisk2018. On both datasets, precision decreases as additional symptoms are included, indicating that lower-ranked symptoms are less consistently supported by the posts. These findings provide evidence of the correspondence between PAT's symptom explanations and users' expressed experiences, while highlighting the need for caution when interpreting lower-ranked symptoms.
\begin{table*}[tp]
\centering
\small
\caption{Human evaluation of textual support for the top-ranked symptom explanations produced by PAT.}
\label{tab:symptom_eval}
\centering
\begin{tabular}{lcc}
\hline
Evaluation metric & SWDD & eRisk2018 \\
\hline
Top-1 symptom precision & 90.0\% & 61.1\% \\
Top-2 symptom precision & 89.0\% & 57.2\% \\
Top-3 symptom precision & 87.3\% & 52.6\% \\
\hline
\end{tabular}
\end{table*}


\begin{thebibliography}{36}
\providecommand{\natexlab}[1]{#1}

\bibitem[{Agarwal et~al.(2024)Agarwal, Milintsevich, Metivier, Rotharmel, Dias,
  and Dollfus}]{agarwal2024analyzing}
Navneet Agarwal, Kirill Milintsevich, Lucie Metivier, Maud Rotharmel, Ga{\"e}l
  Dias, and Sonia Dollfus. 2024.
\newblock \href {https://aclanthology.org/2024.lrec-main.87/} {Analyzing
  symptom-based depression level estimation through the prism of psychiatric
  expertise}.
\newblock In \emph{Proceedings of the 2024 Joint International Conference on
  Computational Linguistics, Language Resources and Evaluation (LREC-COLING
  2024)}, pages 974--983, Torino, Italia. ELRA and ICCL.

\bibitem[{Ahmed et~al.(2022)Ahmed, Srivastava, Yun, and Lin}]{ahmed2022eandc}
Usman Ahmed, Gautam Srivastava, Unil Yun, and Jerry Chun-Wei Lin. 2022.
\newblock Eandc: An explainable attention network based deep adaptive
  clustering model for mental health treatment.
\newblock \emph{Future Generation Computer Systems}, 130:106--113.

\bibitem[{Beck(1996)}]{beck1996manual}
Aaron~T Beck. 1996.
\newblock Manual for the beck depression inventory-ii.
\newblock \emph{(No Title)}.

\bibitem[{Cai et~al.(2023)Cai, Wang, Ye, Jin, and Gao}]{cai2023depression}
Yicheng Cai, Haizhou Wang, Huali Ye, Yanwen Jin, and Wei Gao. 2023.
\newblock Depression detection on online social network with multivariate time
  series feature of user depressive symptoms.
\newblock \emph{Expert Systems with Applications}, 217:119538.

\bibitem[{Costa and McCrae(2008)}]{costa2008revised}
Paul~T Costa and Robert~R McCrae. 2008.
\newblock The revised neo personality inventory (neo-pi-r).
\newblock \emph{The SAGE handbook of personality theory and assessment},
  2(2):179--198.

\bibitem[{Devlin et~al.(2019)Devlin, Chang, Lee, and
  Toutanova}]{devlin2019bert}
Jacob Devlin, Ming-Wei Chang, Kenton Lee, and Kristina Toutanova. 2019.
\newblock Bert: Pre-training of deep bidirectional transformers for language
  understanding.
\newblock In \emph{Proceedings of the 2019 conference of the North American
  chapter of the association for computational linguistics: human language
  technologies, volume 1 (long and short papers)}, pages 4171--4186.

\bibitem[{Ding et~al.(2020)Ding, Chen, Fu, and Zhong}]{ding2020depression}
Yan Ding, Xuemei Chen, Qiming Fu, and Shan Zhong. 2020.
\newblock A depression recognition method for college students using deep
  integrated support vector algorithm.
\newblock \emph{IEEE access}, 8:75616--75629.

\bibitem[{Gamaarachchige and Inkpen(2019)}]{gamaarachchige2019multi}
Prasadith~Kirinde Gamaarachchige and Diana Inkpen. 2019.
\newblock Multi-task, multi-channel, multi-input learning for mental illness
  detection using social media text.
\newblock In \emph{Proceedings of the tenth international workshop on health
  text mining and information analysis (LOUHI 2019)}, pages 54--64.

\bibitem[{Hossain et~al.(2021)Hossain, Talukder, and Jahan}]{hossain2021social}
Md~Tazmim Hossain, Md~Arafat~Rahman Talukder, and Nusrat Jahan. 2021.
\newblock Social networking sites data analysis using nlp and ml to predict
  depression.
\newblock In \emph{2021 12th International Conference on Computing
  Communication and Networking Technologies (ICCCNT)}, pages 1--5. IEEE.

\bibitem[{Ji et~al.(2022)Ji, Zhang, Ansari, Fu, Tiwari, and
  Cambria}]{ji2022mentalbert}
Shaoxiong Ji, Tianlin Zhang, Luna Ansari, Jie Fu, Prayag Tiwari, and Erik
  Cambria. 2022.
\newblock Mentalbert: Publicly available pretrained language models for mental
  healthcare.
\newblock In \emph{proceedings of the thirteenth language resources and
  evaluation conference}, pages 7184--7190.

\bibitem[{Jung and Beebe(2016)}]{jung2016psychological}
Carl Jung and John Beebe. 2016.
\newblock \emph{Psychological types}.
\newblock Routledge.

\bibitem[{Kour and Gupta(2022)}]{kour2022hybrid}
Harnain Kour and Manoj~K Gupta. 2022.
\newblock An hybrid deep learning approach for depression prediction from user
  tweets using feature-rich cnn and bi-directional lstm.
\newblock \emph{Multimedia Tools and Applications}, 81(17):23649--23685.

\bibitem[{Koutsoumpis et~al.(2022)Koutsoumpis, Oostrom, Holtrop, Van~Breda,
  Ghassemi, and de~Vries}]{koutsoumpis2022kernel}
Antonis Koutsoumpis, Janneke~K Oostrom, Djurre Holtrop, Ward Van~Breda, Sina
  Ghassemi, and Reinout~E de~Vries. 2022.
\newblock The kernel of truth in text-based personality assessment: A
  meta-analysis of the relations between the big five and the linguistic
  inquiry and word count (liwc).
\newblock \emph{Psychological Bulletin}, 148(11-12):843.

\bibitem[{Lan et~al.(2025)Lan, Han, Cheng, Sheng, Feng, Gao, and
  Li}]{lan2025depression}
Xiaochong Lan, Zhiguang Han, Yiming Cheng, Li~Sheng, Jie Feng, Chen Gao, and
  Yong Li. 2025.
\newblock \href {https://doi.org/10.18653/v1/2025.emnlp-industry.151}
  {Depression detection on social media with large language models}.
\newblock In \emph{Proceedings of the 2025 Conference on Empirical Methods in
  Natural Language Processing: Industry Track}, pages 2155--2171, Suzhou
  (China). Association for Computational Linguistics.

\bibitem[{Lin et~al.(2020)Lin, Hu, Su, Li, Mei, Zhou, and
  Leung}]{lin2020sensemood}
Chenhao Lin, Pengwei Hu, Hui Su, Shaochun Li, Jing Mei, Jie Zhou, and Henry
  Leung. 2020.
\newblock Sensemood: depression detection on social media.
\newblock In \emph{Proceedings of the 2020 international conference on
  multimedia retrieval}, pages 407--411.

\bibitem[{Liu et~al.(2024)Liu, Li, Zhang, Zhang, Wang, Ma, Chen, Yu, and
  Zhang}]{liu2024depression}
Han Liu, Changya Li, Xiaotong Zhang, Feng Zhang, Wei Wang, Fenglong Ma,
  Hongyang Chen, Hong Yu, and Xianchao Zhang. 2024.
\newblock Depression detection via capsule networks with contrastive learning.
\newblock In \emph{Proceedings of the AAAI Conference on Artificial
  Intelligence}, volume~38, pages 22231--22239.

\bibitem[{Losada and Crestani(2016)}]{losada2016test}
David~E Losada and Fabio Crestani. 2016.
\newblock A test collection for research on depression and language use.
\newblock In \emph{International conference of the cross-language evaluation
  forum for European languages}, pages 28--39. Springer.

\bibitem[{Losada et~al.(2018)Losada, Crestani, and
  Parapar}]{losada2018overview}
David~E Losada, Fabio Crestani, and Javier Parapar. 2018.
\newblock Overview of erisk: early risk prediction on the internet.
\newblock In \emph{International conference of the cross-language evaluation
  forum for european languages}, pages 343--361. Springer.

\bibitem[{Narayanan et~al.(2022)Narayanan, Babu, and
  Thandayantavida}]{narayanan2022detection}
S~Ram Narayanan, S~Babu, and Amal Thandayantavida. 2022.
\newblock Detection of depression from social media using deep learning
  approach.
\newblock \emph{Journal of positive school psychology}, 6(4):4909--4915.

\bibitem[{Nguyen et~al.(2022)Nguyen, Yates, Zirikly, Desmet, and
  Cohan}]{nguyen2022improving}
Thong Nguyen, Andrew Yates, Ayah Zirikly, Bart Desmet, and Arman Cohan. 2022.
\newblock \href {https://doi.org/10.18653/v1/2022.acl-long.578} {Improving the
  generalizability of depression detection by leveraging clinical
  questionnaires}.
\newblock In \emph{Proceedings of the 60th Annual Meeting of the Association
  for Computational Linguistics (Volume 1: Long Papers)}, pages 8446--8459,
  Dublin, Ireland. Association for Computational Linguistics.

\bibitem[{Pan and Yang(2009)}]{pan2009survey}
Sinno~Jialin Pan and Qiang Yang. 2009.
\newblock A survey on transfer learning.
\newblock \emph{IEEE Transactions on knowledge and data engineering},
  22(10):1345--1359.

\bibitem[{P{\'e}rez et~al.(2023)P{\'e}rez, Warikoo, Wang, Parapar, and
  Gurevych}]{perez-etal-2023-semantic}
Anxo P{\'e}rez, Neha Warikoo, Kexin Wang, Javier Parapar, and Iryna Gurevych.
  2023.
\newblock \href {https://doi.org/10.18653/v1/2023.emnlp-main.1000} {Semantic
  similarity models for depression severity estimation}.
\newblock In \emph{Proceedings of the 2023 Conference on Empirical Methods in
  Natural Language Processing}, pages 16104--16118, Singapore. Association for
  Computational Linguistics.

\bibitem[{Ravenda et~al.(2025)Ravenda, Bahrainian, Raballo, Mira, and
  Kando}]{ravenda-etal-2025-llms}
Federico Ravenda, Seyed~Ali Bahrainian, Andrea Raballo, Antonietta Mira, and
  Noriko Kando. 2025.
\newblock \href {https://doi.org/10.18653/v1/2025.acl-long.440} {Are {LLM}s
  effective psychological assessors? leveraging adaptive {RAG} for
  interpretable mental health screening through psychometric practice}.
\newblock In \emph{Proceedings of the 63rd Annual Meeting of the Association
  for Computational Linguistics (Volume 1: Long Papers)}, pages 8975--8991,
  Vienna, Austria. Association for Computational Linguistics.

\bibitem[{Sun et~al.(2021)Sun, Wang, Feng, Ding, Pang, Shang, Liu, Chen, Zhao,
  Lu et~al.}]{sun2021ernie}
Yu~Sun, Shuohuan Wang, Shikun Feng, Siyu Ding, Chao Pang, Junyuan Shang,
  Jiaxiang Liu, Xuyi Chen, Yanbin Zhao, Yuxiang Lu, and 1 others. 2021.
\newblock Ernie 3.0: Large-scale knowledge enhanced pre-training for language
  understanding and generation.
\newblock \emph{arXiv preprint arXiv:2107.02137}.

\bibitem[{Tahir et~al.(2025)Tahir, Khalid, Almutairi, Abohashrh, Memon, and
  Khan}]{tahir2025depression}
Waleed~Bin Tahir, Shah Khalid, Sulaiman Almutairi, Mohammed Abohashrh,
  Sufyan~Ali Memon, and Jawad Khan. 2025.
\newblock Depression detection in social media: A comprehensive review of
  machine learning and deep learning techniques.
\newblock \emph{IEEE Access}.

\bibitem[{Vasha et~al.(2023)Vasha, Sharma, Esha, Al~Nahian, and
  Polin}]{vasha2023depression}
Zannatun~Nayem Vasha, Bidyut Sharma, Israt~Jahan Esha, Jabir Al~Nahian, and
  Johora~Akter Polin. 2023.
\newblock Depression detection in social media comments data using machine
  learning algorithms.
\newblock \emph{Bulletin of Electrical Engineering and Informatics},
  12(2):987--996.

\bibitem[{Wang et~al.(2025)Wang, Zi, Sun, Yang, Zhao, and
  Qin}]{wang-etal-2025-end}
Bichen Wang, Yuzhe Zi, Yixin Sun, Hao Yang, Yanyan Zhao, and Bing Qin. 2025.
\newblock \href {https://doi.org/10.18653/v1/2025.emnlp-main.201} {End-to-end
  learnable psychiatric scale guided risky post screening for depression
  detection on social media}.
\newblock In \emph{Proceedings of the 2025 Conference on Empirical Methods in
  Natural Language Processing}, pages 4054--4066, Suzhou, China. Association
  for Computational Linguistics.

\bibitem[{Wang et~al.(2024)Wang, Zi, Zhao, Deng, and Qin}]{wang-etal-2024-esdm}
Bichen Wang, Yuzhe Zi, Yanyan Zhao, Pengfei Deng, and Bing Qin. 2024.
\newblock \href {https://aclanthology.org/2024.lrec-main.556/} {{ESDM}: Early
  {S}ensing depression model in social media streams}.
\newblock In \emph{Proceedings of the 2024 Joint International Conference on
  Computational Linguistics, Language Resources and Evaluation (LREC-COLING
  2024)}, pages 6288--6298, Torino, Italia. ELRA and ICCL.

\bibitem[{Wei et~al.(2022)Wei, Wang, Schuurmans, Bosma, Xia, Chi, Le, Zhou
  et~al.}]{wei2022chain}
Jason Wei, Xuezhi Wang, Dale Schuurmans, Maarten Bosma, Fei Xia, Ed~Chi, Quoc~V
  Le, Denny Zhou, and 1 others. 2022.
\newblock Chain-of-thought prompting elicits reasoning in large language
  models.
\newblock \emph{Advances in neural information processing systems},
  35:24824--24837.

\bibitem[{Wu et~al.(2023)Wu, Wu, Hua, Lin, Zheng, and Yang}]{wu2023exploring}
Jiageng Wu, Xian Wu, Yining Hua, Shixu Lin, Yefeng Zheng, and Jie Yang. 2023.
\newblock Exploring social media for early detection of depression in covid-19
  patients.
\newblock In \emph{Proceedings of the ACM Web Conference 2023}, pages
  3968--3977.

\bibitem[{Yang et~al.(2021)Yang, Li, Ji, Liang, Xie, Tian, Li, and
  Liang}]{yang2021fine}
Tingting Yang, Fei Li, Donghong Ji, Xiaohui Liang, Tian Xie, Shuwan Tian, Bobo
  Li, and Peitong Liang. 2021.
\newblock Fine-grained depression analysis based on chinese micro-blog reviews.
\newblock \emph{Information Processing \& Management}, 58(6):102681.

\bibitem[{Yates et~al.(2017)Yates, Cohan, and Goharian}]{yates2017depression}
Andrew Yates, Arman Cohan, and Nazli Goharian. 2017.
\newblock \href {https://doi.org/10.18653/v1/D17-1322} {Depression and
  self-harm risk assessment in online forums}.
\newblock In \emph{Proceedings of the 2017 Conference on Empirical Methods in
  Natural Language Processing}, pages 2968--2978, Copenhagen, Denmark.
  Association for Computational Linguistics.

\bibitem[{Zhai et~al.(2024)Zhai, Qi, Zhao, Li, Wang, Wang, Yang, and
  Fu}]{zhai2024chinese}
Wei Zhai, Hongzhi Qi, Qing Zhao, Jianqiang Li, Ziqi Wang, Han Wang, Bing Yang,
  and Guanghui Fu. 2024.
\newblock Chinese mentalbert: Domain-adaptive pre-training on social media for
  chinese mental health text analysis.
\newblock In \emph{Findings of the Association for Computational Linguistics:
  ACL 2024}, pages 10574--10585.

\bibitem[{Zhang et~al.(2025)Zhang, Gao, Zhou, and
  He}]{zhang-etal-2025-explainable}
Linhai Zhang, Ziyang Gao, Deyu Zhou, and Yulan He. 2025.
\newblock \href {https://doi.org/10.18653/v1/2025.findings-acl.517}
  {Explainable depression detection in clinical interviews with personalized
  retrieval-augmented generation}.
\newblock In \emph{Findings of the Association for Computational Linguistics:
  ACL 2025}, pages 9927--9944, Vienna, Austria. Association for Computational
  Linguistics.

\bibitem[{Zhang et~al.(2022)Zhang, Chen, Wu, and Zhu}]{zhang2022psychiatric}
Zhiling Zhang, Siyuan Chen, Mengyue Wu, and Kenny~Q Zhu. 2022.
\newblock Psychiatric scale guided risky post screening for early detection of
  depression.
\newblock \emph{arXiv preprint arXiv:2205.09497}.

\bibitem[{Zheng et~al.(2024)Zheng, Guo, and Hong}]{zheng2024cascade}
Tong Zheng, Yanrong Guo, and Richang Hong. 2024.
\newblock Cascade large language model via in-context learning for depression
  detection on chinese social media.
\newblock In \emph{Chinese Conference on Pattern Recognition and Computer
  Vision (PRCV)}, pages 353--366. Springer.

\end{thebibliography}
\end{document}